\documentclass[preprint,12pt]{elsarticle}

\usepackage{amssymb}
\usepackage{amsmath}
\usepackage{pifont}
\usepackage{adjustbox}
\usepackage{booktabs}
\usepackage{multirow}
\usepackage{wrapfig}
\usepackage{caption}
\usepackage{subcaption}
\usepackage[table,dvipsnames]{xcolor}
\usepackage{lineno}

\usepackage{makecell}

\graphicspath{ {./images/} }

\usepackage[pagebackref=true]{hyperref}
\hypersetup{
  colorlinks   = true, %Colours links instead of ugly boxes
  urlcolor     = blue, %Colour for external hyperlinks
  linkcolor    = blue, %Colour of internal links
  citecolor   = blue %Colour of citations
}

\journal{Ecological Informatics}

\begin{document}

\begin{frontmatter}



\title{Data-Efficient Self-Supervised Algorithms for Fine-Grained Birdsong Analysis}


\author[label1]{Houtan Ghaffari\corref{cor1}} %% Author name
\cortext[cor1]{Corresponding author: houtan.ghaffari@ugent.be}
\author[label2]{Lukas Rauch}
\author[label1]{Paul Devos}

\affiliation[label1]{organization={Department of Information Technology, Ghent University},%Department and Organization
            city={Ghent},
            country={Belgium}}

\affiliation[label2]{organization={Intelligent Embedded Systems, University of Kassel},%Department and Organization
            city={Kassel},
            country={Germany}}
            
%% Abstract 250-word limit
\begin{abstract}
Research in bioacoustics, neuroscience, and linguistics often uses birdsong as a proxy to acquire knowledge across diverse areas. This requires audio models to annotate and parse the birdsong. Developing such models requires precise, syllable-level annotated training data. Therefore, automated methods that reduce annotation costs are in demand. This work presents a data-efficient birdsong annotator called Residual Multi-Layer Perceptron Recurrent Neural Network. It then presents a three-stage training pipeline for developing reliable birdsong syllable detectors with minimal annotation. The first stage is self-supervised learning from unlabeled data. Two of the most successful pretraining paradigms are explored, namely, masked prediction and online clustering. The second stage is supervised training with effective data augmentation to produce a robust frame-level syllable detector for each individual. The third stage is a semi-supervised post-training step that refines each individual's model using unlabeled data. The effectiveness of this approach is demonstrated for the Canary song in extreme label-scarcity scenarios. From a signal-processing perspective, the Canary song exhibits one of the most challenging spectro-temporal patterns for algorithmic time-series annotation: rapid vocalizations, brief inter-syllabic intervals, fast and broadband frequency sweeps, and spectrally similar syllables that require fine-grained features to distinguish. Hence, a successful syllable detection algorithm for Canary also establishes a robust baseline for other birds. This methodological generalization is validated in a case study of Bengalese Finch song annotation. Finally, the potential of self-supervised embeddings is assessed for linear probing and unsupervised birdsong analysis.
\end{abstract}


% \begin{highlights}
% \item An expressive, yet data-efficient neural network for both few-shot and self-supervised birdsong analysis.
% \item Adapting the masked prediction and clustering self-supervised methods for fine-grained birdsong analysis.
% \item A three-stage development recipe that minimizes annotation costs for birdsong analysis.
% \item Code, practical guidelines, and future directions for practitioners and researchers.
% \end{highlights}

%% Keywords

\end{frontmatter}

%\linenumbers


%% Labels are used to cross-reference an item using \ref command.
%% Use \subsubsection, \paragraph, \subparagraph commands to 
%% start 3rd, 4th and 5th level sections.
%% Refer following link for more details.
%% https://en.wikibooks.org/wiki/LaTeX/Document_Structure#Sectioning_commands
\section{Introduction}
Song acquisition in birds and speech learning in humans have remarkable cultural and biological similarities~\cite{brainard2002songbirds,berwick2011songs}. As such, birds have been studied as a proxy system to elucidate neural mechanisms underlying sensory-motor learning, plasticity, and neurogenesis~\cite{brainard2002songbirds,mets2019learning}. The knowledge of vocal communication in animal models deepens our understanding of valuable biological and linguistic inquiries~\cite{cohen2022automated,morita2021measuring}, for example, investigating the neurogenetic basis for speech and communication disorders~\cite{burkett2015voice}. Many of these studies require the annotation of large volumes of birdsong recordings at the syllable level~\cite{cohen2022automated}. A general definition of a syllable is a brief and uninterrupted pattern of vocalization separated from other syllables by tiny gaps~\cite{berwick2011songs,cohen2022automated}. The annotation procedure consists of a segmentation to locate the syllables’ temporal boundaries, followed by the expert's label assignment.

Some bird species are more popular than others in these studies because of their intriguing and sophisticated vocal behavior~\cite{PODOS2022R1100,HUETZ2004395,cohen2022automated}. However, they may require intensive annotation efforts due to the complexity of their songs and the diversity of their syllable repertoire~\cite{cohen2022automated}. Notably, automating this procedure is largely independent of the songs' syntactic complexity, e.g, long-distance context dependency~\cite{morita2021measuring}. A popular example is the Canary song, which may have a simple grammar, but exhibits one of the most challenging spectro-temporal patterns for algorithmic time-series annotation. It exhibits rapid, repetitive vocalizations; extremely brief intersyllabic intervals; broadband frequency sweeps in short intervals; and a large repertoire of spectrally similar syllable types that require fine-grained features for discrimination. Hence, basic methods, such as threshold-based segmentation and clustering, are insufficient for precise syllable detection~\cite{houtanfa,cohen2022automated}. As a result, birds like the Canary remain understudied relative to birds with simpler signal patterns, such as the Bengalese Finch~\cite{cohen2022automated}. Also, the difficulty of the annotation may lead to nonreproducible findings~\cite{burkett2015voice,sainburg2020finding,berwick2011songs,boncoraglio2007habitat}. Hence, it is desirable to develop robust, generalizable methods for annotating birdsong.

Deep learning has shown superior performance and automation compared to prior birdsong syllable detectors that rely on classical algorithms~\cite{cohen2022automated,morita2021measuring,houtanfa}. More importantly, a neural network performs segmentation and categorization simultaneously by classifying non-syllable spectrogram frames as background~\cite{cohen2022automated}. However, purely supervised methods do not scale well with the number of birds in this task, because each bird requires a separate model to detect its song patterns, given the idiosyncratic nature of its song and syllable repertoire.

Leveraging unlabeled data via Self-Supervised Learning (SSL) and Semi-Supervised Learning (Semi-SL) methods is gaining attention in computational bioacoustics~\citep{morfi2021deep,rauch2025can,rauch2025unmutepatchtokensrethinking}. These paradigms can address the annotation bottleneck in large-scale birdsong analysis with limited data, particularly for birds with intricate vocal behavior. Moreover, a successful demonstration motivates the publication of large-scale birdsong datasets to develop foundation models for animal communication. Note that this is a different task from bird species classification from outdoor recordings~\cite{hagiwara2023aves,PAPIE}. This work makes the following contributions to fine-grained birdsong analysis:
\begin{itemize}
\item A data-efficient syllable annotation neural network that surpasses the prior art with random initialization in a few-shot setting, while flexible enough to incorporate SSL methods without extensive hyperparameter tuning and large pretraining data.
\item Adapting the masked prediction and clustering SSL methods for fine-grained birdsong analysis.
\item A semi-supervised post-training recipe that improves the results without additional annotation.
\item Detailed experiments and discussion of the results to provide practical guidelines and future directions.
\item Code for utilizing the methods and also expanding it with ease: \url{https://github.com/houtan-ghaffari/BirdsongAnalysisDeepPipeline}
% \textit{a link will be provided after the double-blind review.}
\end{itemize}

\section{Related Works}\label{sec:related_works}

\subsection{Masked Prediction Self-Supervised Learning}\label{sec:SSL_MAE_background}
Devlin~et~al.~\cite{devlin2019bert} introduced BERT, a seminal work in masked prediction SSL that catalyzed the current era of generative AI. Masked language modeling owes its success to at least two major factors. First, language encodes knowledge in an extremely compressed form, already tokenized into discrete semantic pieces that each carry significant information. Although the tokens have sophisticated associations and joint distributions, they are grammatically structured and easy to learn. Second, the Transformer architecture~\cite{vaswani2017attention} is built upon these text-specific properties, which contributed significantly to the success of modern SSL.

These crisp properties are absent in continuous information media because neighboring regions are highly correlated, as in image and audio. He~et~al.~\cite{he2022masked} recognized this bottleneck and leveraged the Vision Transformer (ViT)~\cite{dosovitskiy2021an} to align the architecture with the language domain. Then, they proposed randomly masking 75\% of tokens. This Masked Auro-Encoder (MAE) model demonstrated that using ViT with extreme masking achieves top performance on image benchmarks. Unlike language, a continuous data modality requires a more sophisticated decoder, which they used a lighter ViT for. They showed that a deep decoder reduces the encoder’s role in reconstruction, which allows it to learn high-level features that perform better in downstream classification. They also found that data augmentation is detrimental for MAE. Random cropping and flipping were sufficient, whereas adding color jittering degraded the results, unlike contrastive methods, which rely heavily on augmentation~\cite{grill2020bootstrap,chen2020simple}. It seemed that nullifying the augmentation and masked image modeling were incompatible tasks for a ViT.

Inspired by MAE~\cite{he2022masked}, Huang~et~al.~\cite{huang2022masked} proposed Audio-MAE by applying a similar SSL setup to audio spectrograms, which are akin to grayscale images (albeit superficially in terms of 2D structure). Audio-MAE achieved top performance on six audio and speech classification benchmarks. Similar to MAE, they reported that heavy data augmentation and masked audio modeling were not complementary tasks.

Rauch~et~al.~\cite{rauch2025can} demonstrated that Audio-MAE performs poorly when applied to bird species classification (a different task from birdsong syllable detection), thereby questioning the transferability of general audio dataset pretraining to a finer audio task. They demonstrated that bridging this gap requires not only domain-specific pretraining data but also an adapted finetuning pipeline with proper augmentation techniques. Then, they proposed Bird-MAE, an Audio-MAE for bird species classification, achieving top performance on the BirdSet~\cite{rauch2025_birdset}. This work highlighted the importance of tailoring SSL pipelines for fine-grained audio domains.

\subsection{Clustering Self-Supervised Learning}\label{sec:SSL_Clustering_background}
Caron~et~al.~\cite{caron2018deep} proposed Deep Cluster, an early work that sparked widespread interest in clustering tasks for SSL. It pretrains a convolutional network by alternating between pseudo-labeling the embeddings by k-means and then using them to train the model. This procedure may collapse to a fixed point by overpopulating a single cluster or yield poor representation. It is a type of overfitting that requires regularization. They circumvented collapse by brute-force reassigment of samples to empty clusters at each pseudo-labeling phase, and also weighing the loss function for each category by the inverse of its cluster’s size. This motivates learning a representation with uniform cluster assignment. Two major drawbacks of the Deep Cluster are the complete clustering of the dataset at each epoch and the unprincipled regularization, which does not guarantee adequate representation learning.

Asano~et~al.~\cite{YM.2020Self-labelling} proposed self-labeling via simultaneous clustering and representation learning (SeLA). SeLA improved the SSL field by popularizing the Sinkhorn-Knopp algorithm~\cite{cuturi2013sinkhorn} for label assignment, an iterative fast algorithm for creating doubly stochastic matrices as solutions to the optimal transport problem. This method elegantly applies the equipartition constraint in clustering to prevent collapse. However, SeLa also required a complete pass through the dataset for each pseudo-labeling stage.

Caron~et~al.~\cite{caron2020unsupervised} proposed SwAV, where the model employs a swapped prediction mechanism to predict the cluster code of an augmented view from the representation of another augmented view of the same image. Notably, SwAV uses Sinkhorn-Knopp for online (batch-wise or on-the-fly) clustering to avoid alternating between full-dataset clustering and training at each epoch. They also reported that soft assignment is more effective in online clustering.

A seminal work by Caron~et~al.~\cite{caron2021emerging} proposed knowledge distillation with no labels (DINO), an extension of BYOL~\cite{grill2020bootstrap} and SwAV~\cite{caron2020unsupervised}. DINO combined clustering and ViT, revealing interesting emerging properties in image semantic segmentation. They leveraged the momentum encoder method~\cite{Lillicrap2016,tarvainen2017mean,he2020momentum}, an Exponential Moving Average (EMA) of the model that generates pseudo-targets online. They used the cls-token for clustering without directly leveraging patch tokens, but subsequent work has shown the benefits of using the ViT's patch tokens in SSL~\citep{zhou2021ibot,oquab2024dinov}.

Assran~et~al.~\cite{assran2022masked} extended DINO using masked siamese networks~\cite{NIPS1993_288cc0ff,chen2021exploring}. They applied token masking as an additional augmentation, similar to MAE~\cite{he2022masked}, but limited the task to clustering the cls-token, as in DINO~\cite{caron2021emerging}. Notably, alongside the Sinkhorn-Knopp method, they incorporated mean-entropy maximization~\cite{joulin2012convex} to enhance clustering regularization.

\subsection{Semi-Supervised Learning}
Semi-SL methods learn high-level semantic structures from unlabeled data and rely on the annotations only for learning the fine-grained details of a task~\cite{murphy2022probabilistic}. It involves entropy minimization, either directly as a measure of class overlap~\cite{grandvalet2004semi} or via pseudo-labeling~\cite{lee2013pseudo}, to train the model using its predictions on unlabeled data alongside human-labeled data. However, this rudimentary self-training approach is prone to reinforcing model errors, i.e., confirmation bias. Thus, recent methods leverage unlabeled examples that the model can predict with high confidence to mitigate confirmation bias~\cite{sohn2020fixmatch}. Additionally, the teacher-student framework~\cite{hinton2015distilling} has been remarkably successful across both Semi-SL~\cite{tarvainen2017mean} and SSL~\cite{he2020momentum,grill2020bootstrap,caron2021emerging}. Particularly, Tarvainen~et~al.~\cite{tarvainen2017mean} proposed the Mean Teacher algorithm, where the pseudo-label of unlabeled data is produced online via an EMA of the model. This method is leveraged to post-train the birdsong syllable detectors in this work.

\subsection{Birdsong Analysis and Position of this Paper}
The theory of sexual selection has been tested extensively using the complexity of birdsong as a model for trait elaboration~\cite{soma2011rethinking}. Neuroscience and linguistics studies of brain activity and behavioral patterns rely on analyses of birdsong structure and similarity across individuals~\cite{terpstra2004analysis,williams2004birdsong,lipkind2011quantification,morita2021measuring}. Researchers have investigated the effect of habitat structure on the evolution of birdsong~\cite{boncoraglio2007habitat}. A remarkable study by Sober and Brainard~\cite{sober2009adult} analyzed birdsong and found that lifelong error correction is not specific to humans but is a general principle of learned vocal behavior. Some studies also investigate linguistic parallels between birdsong and spoken language~\cite{berwick2011songs}.

These works highlight the importance of a robust algorithm for reproducible and accurate birdsong analysis as a prerequisite for such studies~\cite{daou2012computational,cohen2022automated}. A scalable birdsong syllable detector that minimizes annotation cost without compromising reliability will significantly increase the fidelity of downstream studies by narrowing the sources of variability and bias to other experimental and biological factors of interest.

To this end, Cohen~et~al.~\cite{cohen2022automated} proposed TweetyNet, a Convolutional Recurrent Neural Network (CRNN) for syllable detection that works well with moderate amounts of labeled data. This work proposes \textit{Residual Multi-Layer Perceptron Recurrent Neural Network} (Res-MLP-RNN), an alternative model to TweetyNet that is more data-efficient and better suited for the task. It also provides a three-stage framework that leverages unlabeled data to minimize annotation costs algorithmically. Masked prediction and clustering SSL methods have demonstrated strong performance across various data modalities. However, finding suitable SSL settings for modest model sizes and limited data is challenging and underexplored in fine-grained birdsong analysis. Combined with Res-MLP-RNN, which replaces data-hungry Transformers, this work presents two successful SSL approaches for learning semantic features of birdsong structure. It also integrates best practices for a post-training Semi-SL recipe to further refine the models without annotation costs.

\section{Methods}\label{sec:methods}
\begin{figure}[t!]
    \centering
    \includegraphics[width=\textwidth]{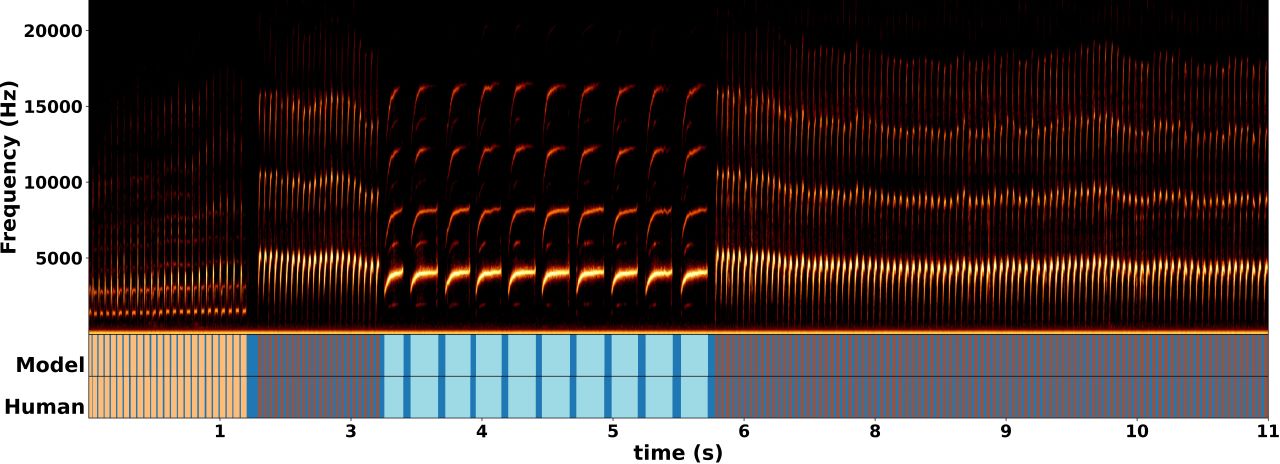}
    \hfill
    \caption{An example of syllable detection using the proposed model and the three-stage training framework with MAE pretraining. There are nearly eight thousand classified frames in this one short excerpt of a Canary song.}
    \label{fig:prediction}
\end{figure}

This section describes the dataset and its utilization in this work. Next, it introduces the Res-MLP-RNN architecture and its application to different tasks. Informed by the model and dataset, it describes two SSL algorithms for pretraining birdsong models. Afterward, it explains the birdsong syllable detection task, using standard supervised training and a follow-up Semi-SL post-training. \autoref{fig:prediction} illustrates an example of a desired output in birdsong syllable detection. The proposed training framework has three clear stages: (i) SSL pretraining using all available birdsong datasets, (ii) individual-specific models for each bird via supervised syllable detection, (iii) Semi-SL post-training to refine the models with both labeled and unlabeled data of each bird.

% Data Acquisition and Split
\subsection{Data: Acquisition, Split, Preprocessing, Augmentation, and Notation}\label{sec:data}
\textbf{Data acquisition}: The dataset consists of laboratory recordings from three Canaries, open-sourced by Cohen~et~al.~\cite{cohen2022automated}. They are sampled at 44.1~kHz and fully annotated at the syllable level. The dataset is utilized carefully to simulate challenging situations and provide an estimate of the annotation required in practice. This evaluates the method's potential to reduce annotation costs for reliable large-scale studies.

\textbf{Data split}: These datasets are highly imbalanced, as some syllables occur only a few times, whereas others are vocalized thousands of times. Hence, the dataset split must be done with care and purpose. First, a minimal few-shot set is extracted for each bird, comprising $\sim 0.5\%$ of the dataset. A `shot' refers to a recording, which contains one or more songs, and each song has a variable number of syllable types with drastically different numbers of repetitions. Notice that not all syllable and song types are present in a single recording. Additionally, the recordings vary in duration, ranging from 1 to 40 seconds. This few-shot set is the minimum number of recording files required to ensure that each syllable is vocalized at least once. The remaining data is split into training and test sets, comprising $\sim 2\%$ and $\sim 98\%$ of the dataset, respectively. Some experiments only leverage the few-shot set for training, while others add half or all of the training data to this few-shot set, indicated by $+1\%$ and $+2\%$, respectively. \autoref{tab:data} shows the details of the dataset split. The SSL methods use the full dataset from all birds together (without labels). The Semi-SL post-training is individual-specific and uses the target bird's test set as the unlabeled dataset. No validation set is used in any experiment, as the aim is to demonstrate the robustness and potential of the proposed method in a label-scarce scenario without additional guidance.

% \begin{wraptable}{r}{0.5\textwidth}
% \vspace{-10pt}
\begin{table}
\setlength\tabcolsep{0pt}
 \centering
 \small
 \caption{The columns show the number of recording files and their total duration in minutes (m) in each split.}
 \begin{tabular*}{\textwidth}{@{\extracolsep{\fill}} l*{5}{c}}
    \toprule
    % \multirow{2}{*}{Bird} & \multirow{2}{*}{Syllables}
     & \multicolumn{3}{c}{Train} & Test & \\
     \cline{2-4} \cline{5-5}
     Bird & Few-shot & +1\% & +2\% & 98\%  & Syllables\\
    \midrule
    llb3 & 13 (3 m) & 39 (10 m) & 65 (14 m) & 2590 (492 m) & 20\\
    llb11 & 10 (3 m) & 30 (12 m) & 50 (19 m) & 1981 (836 m) & 27\\
    llb16 & 13 (3 m) & 27 (8 m) & 41 (13 m) & 1411 (479 m) & 30\\
    \bottomrule
 \end{tabular*}
 \label{tab:data}
\end{table}

\textbf{Handling variable duration}: The variable recording durations prevent efficient utilization of the GPU. Therefore, in each training iteration, a random window is cropped from each recording. A caveat is that syllables do not occur randomly in birdsong, and their temporal order exhibits a syntactic structure that the model should capture. Hence, a large 10-second cropping window is used for the main task of syllable detection, in both supervised training and Semi-SL post-training. At inference time, the test recordings are processed in full length without cropping, allowing the models to leverage the learned structural dependencies among syllables within song patterns. SSL methods are less dependent on a song's full structure. Hence, a shorter 3-second cropping window is used to facilitate the experiments. The recordings shorter than the prespecified window sizes are zero-padded to enable batch processing during training.

\textbf{Preprocessing and features}: The proposed model processes spectrograms with high temporal resolution. A recording waveform is transformed into a power spectrogram using a centered fft window of 512 samples ($11.5$~ms) and a hop length of 64 ($1.5$~ms). The spectrograms were compressed to the decibel scale, and further min-max normalized to $[0, 1]$ range~\cite{GHAFFARI2024102573}.

\textbf{Augmentation}: Suitable data augmentation depends on the characteristics of the task and dataset~\cite{rauch2025can,ferreira2022survey}. For example, pitch-shifting is a common technique in sound event classification~\cite{GHAFFARI2024102573,ferreira2022survey}. However, this task focuses on individual birds with controlled lab recordings. Hence, there is no noticeable deviation in the frequency components of the vocalizations from rendition to rendition. Pitch-shifting is useful when multiple sources produce similar sound events. Examples include species classification or speech processing when the dataset contains individuals of different age groups with diverse physiological characteristics. The following describes promising techniques that consider the characteristics of the task and dataset:
\begin{itemize}
    \item \textbf{Random Gain (RG)}: modulates the input's relative amplitude by $\alpha~\sim~\mathcal{U}(0.5, 1)$. This is a reasonable technique because birds adjust their vocal volume in different communication contexts~\cite{ZOLLINGER2015289}. It also simulates factors such as bird-microphone distance and orientation.
    \item \textbf{Color Noise (CN)}: adds band-specific background noise to the waveform with $SNR \sim \mathcal{U}(5, 30)$ decibels. The procedure consists of generating a white (Gaussian) noise, and modifying its spectral profile with a frequency mask from the $1/f^{\beta}$ power-law, where $f$ denotes each unit of frequency bandwidth, and $\beta \sim \mathcal{U}(-2, 2)$ is the frequency decay parameter. This is an effective augmentation of the content without disrupting the song structure. It is particularly valuable for SSL pretraining. It also simulates faulty recording or accidental noise in the recording environment, e.g., flapping wings in the cage.
    \item \textbf{Bernoulli Noise (BN)}: zeros out a random portion $p \sim \mathcal{U}(0, 0.3)$ of the spectrogram bins. This is also a useful technique for manipulating content without corrupting the song's temporal and harmonic structure. It also simulates faulty recordings and is useful for model regularization, analogous to input dropout (but without scaling).
    \item \textbf{Time-Frequency Masking (TF-Mask)}: zeros out a random number of contiguous time and frequency stripes~\cite{Park_2019}. The temporal mask length ranges from 0 to 105 ms, and the frequency mask length ranges from 0 to 700 Hz. This technique corrupts the temporal and harmonic structure of the songs. If adequate computational resources (GPU) are available to train for many epochs (to see many corrupted versions of the same song) with a large cropping window (to capture song patterns better) on a moderate training size (to include diverse song patterns), TF-Mask potentially benefits the model as it forces it to fill the gaps. Its utility in this low-resource setup will be tested in ablation studies.
\end{itemize}

\textbf{Notation}: A batch of input is denoted by $x \in \mathbb{R}^{B \times T \times F}$, where $B$ is the batch size, $F$ is the number of frequency bins, and $T$ indicates the number of time frames. A corresponding batch of labels is denoted by $y \in \{1, \dots, C\}^{B \times T}$, which is a sequence of $C-1$ individual-specific syllable categories plus the background class for $T$ time frames. The background is assigned to originally unlabeled time frames~\cite{cohen2022automated}. In SSL parlance, a modified version of an input (e.g., augmented or masked) is called a `view'. This work adopts this terminology to specify the view and denotes it by $\Tilde{x}$. If it is necessary to mention more than one view of the same input, they are distinguished by a superscript, e.g., $\Tilde{x}^{(1)}$ and $\Tilde{x}^{(2)}$ denote two views of $x$.

\subsection{Residual Multilayer-Perceptron Recurrent Neural Network and Tasks}\label{sec:models}
Despite the strong inductive bias of convolutional networks toward capturing local patterns~\cite{he2016deep,caron2020unsupervised,woo2023convnext}, this is not the best property for this fine-grained task. We must model sophisticated harmonic and temporal dependencies at variable distances while preserving the temporal dimension. It is better not to irreversibly mix adjacent frames via pooling. Modern audio and vision models, particularly MAE~\cite{he2022masked} and clustering~\cite{caron2021emerging} SSL, rely on large variants of Vision Transformer (ViT)~\cite{dosovitskiy2021an}. However, they require large pretraining datasets, whereas available birdsong datasets are relatively small compared with those in other fields. More importantly, a self-attention module is computationally prohibitive here because a minute-long recording can yield tens of thousands of frames. Moreover, a drawback of pretrained audio Transformers is that the formulation is too general. Although they provide strong initialization for finetuning in downstream tasks, they perform poorly on linear probing~\cite{rauch2025can,he2022masked,huang2022masked}. This indicates that the pretrained model lacks sufficient semantic understanding of the dataset, leading to failure on more fine-grained tasks within the same data modality, as demonstrated by Rauch~et~al.~\cite{rauch2025can}. We need a model with strong inductive biases for this task to handle small annotation while remaining sufficiently flexible for successful SSL pretraining at scales determined by data availability.

\begin{figure}[t!]
    \centering
    \includegraphics[width=\textwidth]{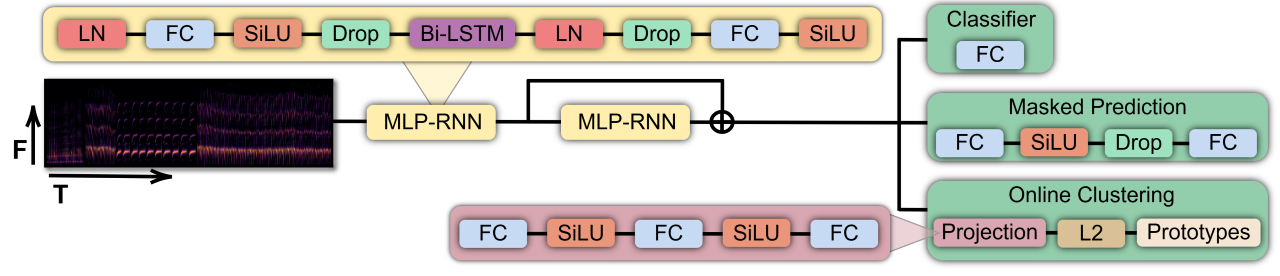}
    \caption{The proposed Res-MLP-RNN neural network architecture with an input spectrogram example. The Masked Prediction and Online Clustering heads are used in two distinct SSL tasks (not multitasking). The Classifier head is used for supervised and post-training Semi-SL syllable detection tasks. FC is Fully-Connected; LN is LayerNorm; Drop is Dropout; Prototypes are learnable cluster centers.}
    \label{fig:architecture}
    % \vspace{-10pt}
\end{figure}

\autoref{fig:architecture} illustrates the proposed architecture that alleviates the stated challenges. The design is partly inspired by the Transformer~\cite{vaswani2017attention} and is optimized for memory and data efficiency for this fine-grained task. It combines layer normalization~\cite{ba2016layer}, linear projections with Sigmoid Linear Unit (SiLU) activation~\cite{ramachandran2017swish}, bi-directional Long Short-Term Memory (LSTM)~\cite{lstm} layers, and dropouts~\cite{srivastava2014dropout}. The model uses two MLP-RNN blocks with a residual connection~\cite{he2016deep}, which is effective at capturing the harmonic and temporal structures. This model outperforms the best lightweight prior model for the task~\cite{cohen2022automated} in a few-shot setting. The appropriate inductive biases largely prevent overfitting, and the model can be trained without a validation set because there is no significant domain shift here. Meanwhile, it is flexible enough to increase its capacity to learn powerful representations from a small unlabeled dataset and to support diverse SSL objectives, without extensive hyperparameter tuning.

The part of the model before the task heads is referred to as the encoder. The masked prediction head is for the birdsong MAE algorithm. Unlike prior MAE works, this model does not require a sophisticated encoder-decoder design. The online clustering head is for the Online Syllable Clustering (OSC) algorithm. This model learns an invariant representation under heavy data augmentation, a requirement for OSC. The classification head is used for downstream syllable detection, where the model is either initialized randomly or leverages the encoder’s knowledge from an SSL task.

The model is denoted by $f_{\theta}$, where $\theta$ indicates parameters. The OSC pretraining and the Semi-SL post-training algorithms require an extra momentum encoder model~\cite{tarvainen2017mean,he2020momentum}, which is an EMA of the main model with an identical architecture. This work refers to the main model as the student and to its EMA version as the teacher. The teacher is denoted as $f_{\theta'}$. Regardless of the algorithm, the student is trained via backpropagation with some objective function. However, at training batch-step $j$, the teacher is updated via the EMA of the student's parameters,
\begin{equation}\label{eq:teacher_update_rule}
\theta'_j = \lambda\theta'_{j-1} + (1 - \lambda) \theta_j,
\end{equation}
where $\lambda \in [0, 1]$ is the decay factor that determines how quickly the teacher incorporates the student's recent state. Only the student is kept after the training.

\subsection{Self-Supervised Algorithm 1: Birdsong Masked Syllable Modeling}\label{sec:ssl_mae}
This section details the procedure for developing the birdsong MAE. At each training iteration, an augmented spectrogram view $\Tilde{x}$ is created for each input $x$ by applying gain modulation and color noise. For every block of 200 frames in $\Tilde{x}$, a random starting position is selected within the first 100 frames of that block, and a random window of 50 to 200 consecutive frames from that starting point is masked to zero. The task is to reconstruct $x$ from $\Tilde{x}$ by nullifying the augmentation and masking. Denote the model prediction by $z=f_{\theta}(\Tilde{x})$, where $z \in \mathbb{R}^{B \times T \times F}$. The following objective function is minimized with respect to the $\theta$,
\begin{equation}\label{eq:supervised_loss}
\ell_{mae}(z, x) = \frac{1}{BTF} \sum_{b=1}^{B} ||z_b - x_b||_2^2.
\end{equation}

\begin{figure*}[!t]
    \centering
    \includegraphics[width=\textwidth]{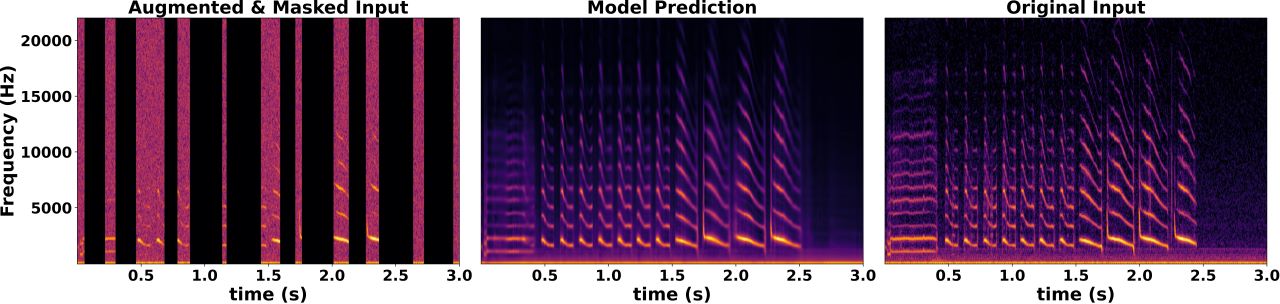}
    \hfill
    \caption{An example from the birdsong MAE model for the masked prediction task.}
    \label{fig:rec_task}
\end{figure*}

The model was trained for 200 epochs using the Adam optimizer~\citep{kingma2014adam}, a linear learning rate warmup from $1\mathrm{e}{-5}$ to $1\mathrm{e}{-3}$ in 20 epochs, then, a constant rate for 10 epochs, and finally, a cosine decay for the remaining 170 epochs to the minimum learning rate of $1\mathrm{e}{-5}$. The experiment was conducted once, and the resulting SSL model was used in all subsequent experiments. \autoref{fig:rec_task} shows an example of an input-prediction-target for this task.

\subsection{Self-Supervised Algorithm 2: Birdsong Online Syllable Clustering}\label{sec:ssl_osc}
This SSL algorithm utilizes the teacher-student framework. The teacher is initialized as an exact copy of the student and is updated according to Eq.~(\ref{eq:teacher_update_rule}), with the decay factor increasing linearly from 0.995 to 0.99998 during the first half of training and held constant thereafter. At each training iteration, two augmented spectrogram views $\Tilde{x}^{(1)}$ and $\Tilde{x}^{(2)}$ are created for each input by applying gain modulation, color noise, and Bernoulli noise. Denote the student's predictions by $z^{(1)}=f_{\theta}(\Tilde{x}^{(1)})$ and $z^{(2)}=f_{\theta}(\Tilde{x}^{(2)})$. Correspondingly, denote the teacher's predictions by $z'^{(1)}=f_{\theta'}(\Tilde{x}^{(1)})$ and $z'^{(2)}=f_{\theta'}(\Tilde{x}^{(2)})$. All predictions for a batch of input are in $\mathbb{R}^{B \times T \times K}$, where $K=1024$ is the number of clusters (prototype vectors). The outputs are converted into probabilities by softmax,
\begin{equation}\label{eq:temp_softmax}
    p^*_{b,t,i} = \frac{\exp{(z^*_{b,t,i} / \tau^*)}}{\sum_{k=1}^{K} \exp{(z^*_{b,t,k} / \tau^*)}},
\end{equation}
where $b$ indexes the batch axis, $t$ indexes the time axis, $i$ indexes the output categories, and $*$ is a placeholder for either view from the student or teacher. The temperatures for student and teacher are $\tau=0.1$ and $\tau'=0.04$, respectively. This sharpening is critical to prevent uninformative clustering. Because the teacher's assignments are regularized via the Sinkhorn-Knopp algorithm~\cite{caron2020unsupervised,YM.2020Self-labelling,cuturi2013sinkhorn} to promote diverse cluster assignments and prevent collapse. The teacher assignments are used as targets for the following swapped-view cross-entropy loss~\cite{caron2020unsupervised,caron2021emerging},
\begin{equation}
    \ell_{ce} = L(p^{(1)},p'^{(2)}) + L(p^{(2)},p'^{(1)}),
\end{equation}
where,
\begin{equation}
L(p^{(1)},p'^{(2)}) = -\frac{1}{BT} \sum_{b,t,k} p'^{(2)}_{b,t,k} \log{(p^{(1)}_{b,t,k})}.
\end{equation}

\begin{figure}%{r}{0.5\textwidth}
    % \vspace{-12pt}
    \centering
    \includegraphics[width=0.8\textwidth]{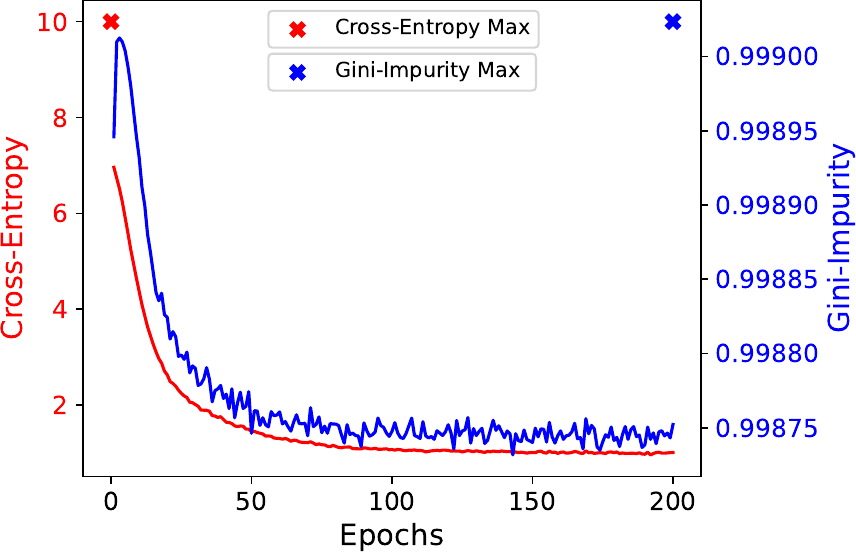}
    \caption{Online Syllable Clustering loss. Maximizing the Gini Impurity enforces the equipartition constraint while allowing the cross-entropy loss to be minimized gradually.}
    \label{fig:ssl_oc_loss}
    % \vspace{-15pt}
\end{figure}

Assran~et~al.~\cite{assran2022masked} showed that maximizing the mean-entropy~\cite{joulin2012convex} of the student's predictions is an effective regularizer that improves clustering SSL. The underlying mechanism of the Sinkhorn-Knopp and mean-entropy maximization is to promote uniform cluster assignments to prevent collapse. This work proposes to maximize the Gini impurity~\cite{breiman2017classification} instead. It has the same effect but with the advantage of being a quadratic function rather than a logarithmic one. Thus, it has better numerical stability for backpropagation. The mean Gini impurity loss is the sum-square of the average student's predicted probabilities across the batch, frames, and views,
\begin{equation}
    \ell_{gini}(\bar{p}) = 1 - \sum_{k=1}^{K} \bar{p}_k^2,
\end{equation}
where,
\begin{equation}\label{eq:gini_average}
    \bar{p} = \frac{1}{2BT}\left( \sum_{b,t} p^{(1)}_{b,t} + \sum_{b,t} p^{(2)}_{b,t}\right),
\end{equation}
and the vectorized summation on the category axis is implicit in Eq.~(\ref{eq:gini_average}). Do not confuse the view index (in parentheses) with the square exponent. Putting everything together, the following objective function is minimized with respect to the student's parameters,
\begin{equation}
    \ell_{osc} = \ell_{ce} - \ell_{gini}.
\end{equation}

The model is trained for 200 epochs using the AdamW optimizer~\cite{loshchilov2018decoupled}, weight decay of $5\mathrm{e}{-2}$, linear learning rate warmup from $1\mathrm{e}{-6}$ to $5\mathrm{e}{-4}$ in 20 epochs, a constant rate for 10 epochs, and cosine decay for the remaining 170 epochs to the minimum of $1\mathrm{e}{-6}$. \autoref{fig:ssl_oc_loss} shows the pretraining loss profile, which indicates no sign of degeneracy for either loss function.

\subsection{Birdsong Syllable Detection}\label{sec:main_task_supervised}
In all experiments and ablations, the model is trained for 300 epochs using the AdamW optimizer~\cite{loshchilov2018decoupled}, a weight decay of $1\mathrm{e}{-3}$, a linear learning rate warmup from $1\mathrm{e}{-6}$ to $1\mathrm{e}{-3}$ in 30 epochs, and cosine decay for the remaining 270 epochs to the minimum of $1\mathrm{e}{-6}$. Denote the model's prediction by $z=f_{\theta}(x)$. The outputs are transformed to probabilities $p \in \mathbb{R}^{B \times T \times C}$ using Eq.(\ref{eq:temp_softmax}) with $\tau=1$. The following cross-entropy loss is minimized with respect to $\theta$ by using labels $y$,
\begin{equation}\label{eq:supervised_ce}
    \ell_{ce} = -\frac{1}{BT}\sum_{b,t} log(p_{b,t,c}), \qquad \textrm{where} \quad c=y_{b,t}.
\end{equation}

\subsubsection{Semi-Supervised Post-Training}\label{sec:main_task_semisupervised}
This stage employs a teacher-student framework for Semi-SL post-training to potentially refine each individual's model using its labeled and unlabeled datasets~\cite{tarvainen2017mean,houtanfa}. The teacher and student models are initialized with the parameters after the supervised stage. The teacher is updated using Eq.~(\ref{eq:teacher_update_rule}), with the decay factor increasing linearly from 0.995 to 0.99998 during the first quarter of training and fixed thereafter. While Eq.~(\ref{eq:supervised_ce}) is used for the labeled data, a consistency loss simultaneously distills unlabeled data by bootstrapping the model. Denote the teacher's prediction for unlabeled input by $z' = f_{\theta'}(x)$. The outputs are transformed to probabilities $p' \in \mathbb{R}^{B \times T \times C}$ using Eq.~(\ref{eq:temp_softmax}) with $\tau'=1$. These probabilities are further converted to hard pseudo-labels $y' \in  \mathbb{R}^{B \times T}$  by choosing the category with the highest probability. An augmented view $\Tilde{x}$ is created for each input $x$ by applying gain modulation, color noise, and Bernoulli noise. The student should match the pseudo-labels by predicting $\Tilde{x}$. Denote the student’s prediction by $z = f_{\theta}(\Tilde{x})$, which are transformed to probabilities $p \in \mathbb{R}^{B \times T \times C}$ using Eq.~(\ref{eq:temp_softmax}) with $\tau=1$. The following consistency loss enables semi-supervised training,
\begin{equation}\label{eq:consistency_loss}
    \ell_{u} = -\frac{1}{BT}\sum_{b,t} log(p_{b,t,c'}), \qquad \textrm{where} \quad c'=y'_{b,t}.
\end{equation}

To avoid the confirmation bias (enforcing wrong predictions), only output time frames with the teacher’s predicted probability above 0.95 are allowed to participate in Eq.~(\ref{eq:consistency_loss}). The student is trained via backpropagation by combining the two loss functions in Eq.~(\ref{eq:supervised_ce}) and Eq.~(\ref{eq:consistency_loss}). The student is trained for 30 epochs using the AdamW optimizer~\cite{loshchilov2018decoupled}, a weight decay of $1\mathrm{e}{-3}$, a linear learning rate warmup from $1\mathrm{e}{-7}$ to $1\mathrm{e}{-4}$ in 3 epochs, and cosine decay for 27 epochs to the minimum learning rate of $1\mathrm{e}{-7}$.

\begin{table}[t!]
\setlength\tabcolsep{0pt}
 \centering
 \scriptsize
 \caption{Res-MLP-RNN (RMR) outperforms TweetyNet using random initialization and without augmentation across multiple train set sizes. The numbers after the RMR indicate the model's hidden dimension. The red entries are the ones that are lower than TweetyNet. RMR-32 is approximately $24\times$ smaller than TweetyNet while still achieving competitive performance. The first column shows that the RMR is more robust against overfitting, even with nearly $9\times$ more parameters than TweetyNet.}
 \begin{tabular*}{\columnwidth}{@{\extracolsep{\fill}} l*{8}{c}}
    \toprule
    \multirow{2}{*}{Model} & \multirow{2}{*}{P} & \multirow{2}{*}{Bird} & \multicolumn{2}{c}{Few-shot} & \multicolumn{2}{c}{Few-shot + 1\%} & \multicolumn{2}{c}{Few-shot + 2\%}\\
    \cline{4-5} \cline{6-7} \cline{8-9}
    & & & Acc & F1 & Acc & F1 & Acc & F1\\
    \midrule   
    \multirow{3}{*}{TweetyNet} & \multirow{3}{*}{1.1 M} & llb \space 3 & 86.8 $\pm$ 0.8 & 53.1 $\pm$ 0.8 & 94.5 $\pm$ 0.3 & 79.8 $\pm$ 1.2 & 96.1 $\pm$ 0.1 & 86.0 $\pm$ 0.8 \\
    & & llb 11 & 85.1 $\pm$ 1.2 & 35.3 $\pm$ 4.5 & 96.6 $\pm$ 0.4 & 79.0 $\pm$ 4.0 & 97.7 $\pm$ 0.1 & 86.9 $\pm$ 0.6 \\
    & & llb 16 & 92.4 $\pm$ 0.3 & 51.8 $\pm$ 0.8 & 95.0 $\pm$ 0.3 & 66.3 $\pm$ 2.9 & 95.0 $\pm$ 0.2 & 64.9 $\pm$ 1.8 \\
    \midrule
    \multirow{3}{*}{RMR-32} & \multirow{3}{*}{48 k} & llb \space 3 & \textcolor{red}{85.7 $\pm$ 0.9} & 54.8 $\pm$ 2.3 & \textcolor{red}{93.4 $\pm$ 0.3} & \textcolor{red}{73.0 $\pm$ 2.1} & \textcolor{red}{93.8 $\pm$ 0.5} & \textcolor{red}{70.9 $\pm$ 1.7} \\
    & & llb 11 & 86.9 $\pm$ 0.6 & 41.0 $\pm$ 2.0 & \textcolor{red}{95.0 $\pm$ 0.4} & \textcolor{red}{67.1 $\pm$ 2.8} & \textcolor{red}{95.6 $\pm$ 0.4} & \textcolor{red}{63.9 $\pm$ 2.2}\\
    & & llb 16 & \textcolor{red}{90.4 $\pm$ 0.6} & \textcolor{red}{46.6 $\pm$ 3.0} & \textcolor{red}{94.0 $\pm$ 0.2} & \textcolor{red}{55.3 $\pm$ 1.8} & \textcolor{red}{93.4 $\pm$ 0.3} & \textcolor{red}{51.0 $\pm$ 1.6}\\
    \midrule
    \multirow{3}{*}{RMR-64} & \multirow{3}{*}{172 k} & llb \space 3 & 87.7 $\pm$ 0.5 & 57.7 $\pm$ 0.8 & 95.0 $\pm$ 0.1 & 82.0 $\pm$ 0.5 & \textcolor{red}{95.7 $\pm$ 0.1} & \textcolor{red}{84.5 $\pm$ 0.8} \\
    & & llb 11 & 91.2 $\pm$ 0.5 & 59.4 $\pm$ 2.6 & 97.5 $\pm$ 0.0 & 84.5 $\pm$ 0.4 & \textcolor{red}{97.5 $\pm$ 0.1} & \textcolor{red}{83.3 $\pm$ 2.1}\\
    & & llb 16 & 92.4 $\pm$ 0.3 & 56.1 $\pm$ 1.1 & 95.6 $\pm$ 0.1 & 70.9 $\pm$ 1.2 & 95.3 $\pm$ 0.2 & 66.7 $\pm$ 2.9\\
    \midrule
    \multirow{3}{*}{RMR-128} & \multirow{3}{*}{647 k} & llb \space 3 & 88.2 $\pm$ 0.8 & 59.1 $\pm$ 1.3 & 95.5 $\pm$ 0.2 & 83.9 $\pm$ 0.5 & 96.2 $\pm$ 0.1 & 86.5 $\pm$ 0.9 \\
    & & llb 11 & 93.2 $\pm$ 0.6 & 67.7 $\pm$ 2.6 & 97.9 $\pm$ 0.1 & 89.5 $\pm$ 0.8 & 98.0 $\pm$ 0.1 & 89.8 $\pm$ 0.8\\
    & & llb 16 & 93.4 $\pm$ 0.2 & 62.7 $\pm$ 1.3 & 96.4 $\pm$ 0.1 & 81.0 $\pm$ 1.3 & 96.4 $\pm$ 0.1 & 79.3 $\pm$ 0.6\\
    \midrule
    \multirow{3}{*}{RMR-160} & \multirow{3}{*}{998 k} & llb \space 3 & 88.0 $\pm$ 1.2 & 59.5 $\pm$ 2.7 & 95.7 $\pm$ 0.1 & 85.2 $\pm$ 0.4 & 96.2 $\pm$ 0.1 & 86.8 $\pm$ 0.7 \\
    & & llb 11 & 93.1 $\pm$ 0.6 & 66.9 $\pm$ 1.7 & 97.9 $\pm$ 0.0 & 89.0 $\pm$ 0.3 & 98.1 $\pm$ 0.0 & 90.2 $\pm$ 0.3\\
    & & llb 16 & 93.5 $\pm$ 0.2 & 64.0 $\pm$ 1.2 & 96.7 $\pm$ 0.1 & 83.7 $\pm$ 0.9 & 96.6 $\pm$ 0.1 & 81.9 $\pm$ 1.5\\
    \midrule
    \multirow{3}{*}{RMR-192} & \multirow{3}{*}{1.4 M} & llb \space 3 & 88.7 $\pm$ 0.4 & 60.0 $\pm$ 1.6 & 95.8 $\pm$ 0.0 & 85.8 $\pm$ 0.3 & 96.4 $\pm$ 0.1 & 87.3 $\pm$ 0.6 \\
    & & llb 11 & 93.6 $\pm$ 0.4 & 69.5 $\pm$ 0.8 & 98.0 $\pm$ 0.1 & 90.2 $\pm$ 0.5 & 98.2 $\pm$ 0.0 & 90.3 $\pm$ 0.5\\
    & & llb 16 & 93.6 $\pm$ 0.3 & 65.0 $\pm$ 0.9 & 96.7 $\pm$ 0.0 & 84.2 $\pm$ 0.3 & 96.7 $\pm$ 0.1 & 83.3 $\pm$ 1.1\\
    \midrule
    \multirow{3}{*}{RMR-256} & \multirow{3}{*}{2.5 M} & llb \space 3 & \textbf{89.3 $\pm$ 0.1} & 62.3 $\pm$ 1.1 & 95.9 $\pm$ 0.1 & 85.8 $\pm$ 0.3 & 96.5 $\pm$ 0.1 & 88.2 $\pm$ 0.6 \\
    & & llb 11 & 93.5 $\pm$ 0.4 & 69.0 $\pm$ 1.2 & 98.0 $\pm$ 0.1 & 90.3 $\pm$ 0.3 & 98.2 $\pm$ 0.0 & 91.1 $\pm$ 0.6\\
    & & llb 16 & 93.9 $\pm$ 0.2 & 67.7 $\pm$ 0.5 & 96.8 $\pm$ 0.1 & 85.4 $\pm$ 0.7 & 96.9 $\pm$ 0.0 & 85.1 $\pm$ 0.7\\
    \midrule
    \multirow{3}{*}{RMR-512} & \multirow{3}{*}{9.8 M} & llb \space 3 & 89.0 $\pm$ 0.5 & \textbf{62.4 $\pm$ 1.4} & \textbf{96.0 $\pm$ 0.2} & \textbf{86.8 $\pm$ 0.7} & \textbf{96.5 $\pm$ 0.1} & \textbf{88.8 $\pm$ 0.2} \\
    & & llb 11 & \textbf{93.6 $\pm$ 0.4} & \textbf{70.3 $\pm$ 1.5} & \textbf{98.1 $\pm$ 0.0} & \textbf{90.9 $\pm$ 0.3} & \textbf{98.3 $\pm$ 0.0} & \textbf{91.6 $\pm$ 0.5}\\
    & & llb 16 & \textbf{94.2 $\pm$ 0.2} & \textbf{71.3 $\pm$ 0.7} & \textbf{96.9 $\pm$ 0.1} & \textbf{87.1 $\pm$ 0.5} & \textbf{97.0 $\pm$ 0.0} & \textbf{87.0 $\pm$ 0.3}\\
    \bottomrule
 \end{tabular*}
 \label{tab:tweety_vs_ours_canary}
\end{table}

\section{Experiments and Results}\label{sec:results}
The results report the frame-wise accuracy and frame-wise macro-averaged F1-score. The accuracy adequately measures the quality of the output frame predictions and their boundaries. This is important for analyzing syllable characteristics, for example, the distribution of their duration. Birdsong datasets are inherently highly imbalanced because birds vocalize syllables at drastically different rates. Thus, the macro-averaged F1-score provides a good sense of how well different syllables are distinguished. This is important for downstream studies that examine the syntactic aspects of the songs and the sequencing of syllables. The highest scores are always bolded in tables, and, if necessary, the second-highest scores are underlined.

\begin{table}[t]
\setlength\tabcolsep{0pt}
 \centering 
 \small
 \caption{Linear probing results using pretrained SSL models and various training data sizes. Both SSL tasks demonstrate strong semantic understanding of the dataset and task and are well-suited to a low-data regime.}
 \begin{tabular*}{\columnwidth}{@{\extracolsep{\fill}} l*{8}{c}}
    \toprule
    \multirow{2}{*}{Bird} & \multirow{2}{*}{Init} & \multicolumn{2}{c}{Few-shot} & \multicolumn{2}{c}{Few-shot + 1\%} & \multicolumn{2}{c}{Few-shot + 2\%}\\
    \cline{3-4} \cline{5-6} \cline{7-8}
    & & Accuracy & F1 & Accuracy & F1 & Accuracy & F1\\
    \midrule   
    \multirow{2}{*}{llb3}& MAE      & 87.21 & 59.54 & 92.50 & 76.01 & 92.87 & 75.99 \\
                          & OSC  & \textbf{92.31} & \textbf{75.25} & \textbf{95.58} & \textbf{86.11} & \textbf{96.02} & \textbf{87.56} \\
    \midrule
    \multirow{2}{*}{llb11}& MAE     & 87.94 & 51.30 & 94.34 & 73.43 & 94.45 & 71.46 \\
                           & OSC & \textbf{96.04} & \textbf{80.69} & \textbf{97.85} & \textbf{90.30} & \textbf{97.85} & \textbf{90.13} \\
    \midrule
    \multirow{2}{*}{llb16}& MAE     & 90.15 & 52.13 & 93.13 & 64.47 & 93.14 & 63.09 \\
                           & OSC & \textbf{94.24} & \textbf{77.38} & \textbf{96.01} & \textbf{86.37} & \textbf{95.93} & \textbf{85.52} \\
    \bottomrule
 \end{tabular*}
 \label{tab:canary_probe_results}
\end{table}

\subsection{Comparison to Prior Art}\label{sec:rmr_vs_tweety}
This experiment compares the proposed Res-MLP-RNN across seven hidden dimension sizes to TweetyNet~\cite{cohen2022automated}, a lightweight CRNN model for birdsong annotation. The models were initialized randomly and trained on the three training set sizes of each bird without data augmentation. This setup mitigates confounding factors, enabling a fair experiment to assess the utility of each architecture's inductive bias for the task. Each experiment is conducted five times to report the mean and standard deviation. The results are shown in \autoref{tab:tweety_vs_ours_canary}. Despite the Res-MLP-RNN-512 having almost an order of magnitude more parameters, it overfits less and achieves a decent performance on the extremely small few-shot training set (first column). Even the nearly $24\times$ smaller Res-MLP-RNN-32 has comparable results. Note that TweetyNet is a competent model for syllable detection~\cite{cohen2022automated}. All subsequent sections and experiments use the Res-MLP-RNN with 512 latent dimensions.

\subsection{Assessing Self-Supervised Embeddings via Linear Probing}
Linear Probing is a common practice to evaluate the quality of pretrained models. The encoder is frozen, and only the linear classifier is trained without data augmentation. \autoref{tab:canary_probe_results} shows that both MAE and OSC are promising SSL algorithms for birdsong. Remarkably, comparing the few-shot column in \autoref{tab:canary_probe_results} with the results in \autoref{tab:tweety_vs_ours_canary} shows that a frozen OSC is even better than complete finetuning with random initialization.

\begin{table}[t]
\setlength\tabcolsep{0pt}
\centering
\small
\caption{Data augmentation ablation study using llb3 bird with few-shot train size, for pretrained and random initialization.}
\begin{tabular*}{\textwidth}{@{\extracolsep{\fill}} *{10}{c}}
\toprule
\multirow{2}{*}{Gain} & \multirow{2}{*}{Color} & \multirow{2}{*}{Bernoulli} & \multirow{2}{*}{Time-Freq} & \multicolumn{2}{c}{Random-Init} & \multicolumn{2}{c}{MAE-Init} & \multicolumn{2}{c}{OSC-Init}\\
\cline{5-6} \cline{7-8} \cline{9-10}
 & & & & Acc & F1 & Acc & F1 & Acc & F1\\
\midrule												
- & - & - & - & 89.35 & 62.36 & 92.31 & 72.20 & 92.44 & 72.55 \\
\ding{51} & - & - & - & 89.54 & 63.48 & 91.33 & 68.16 & 92.63 & 74.22 \\
- & \ding{51} & - & - & 91.97 & 69.32 & 92.11 & 71.75 & 93.21 & 74.60 \\
- & - & \ding{51} & - & 89.45 & 63.17 & 91.98 & 73.75 & 92.86 & 74.88 \\
- & - & - & \ding{51} & 90.97 & 67.02 & 92.30 & 73.37 & 92.96 & 73.72 \\
\ding{51} & \ding{51} & - & - & \underline{92.42} & \textbf{70.93} & 92.27 & 70.27 & 93.44 & 76.45 \\
\ding{51} & - & \ding{51} & - & 89.37 & 62.54 & 92.65 & 73.89 & 91.50 & 69.98 \\
\ding{51} & - & - & \ding{51} & 90.41 & 66.13 & 92.94 & 74.45 & 93.31 & 74.57 \\
- & \ding{51} & \ding{51} & - & 91.48 & 68.26 & 92.56 & 72.92 & 93.48 & 76.94 \\
- & \ding{51} & - & \ding{51} & \textbf{92.52} & 69.78 & \underline{93.32} & \underline{76.93} & 93.42 & \underline{77.09} \\
- & - & \ding{51} & \ding{51} & 89.85 & 63.41 & 93.02 & 74.99 & 92.88 & 75.25 \\
% \rowcolor{Dandelion}
\ding{51} & \ding{51} & \ding{51} & - & 92.25 & \underline{70.68} & \textbf{93.53} & \textbf{77.57} & 93.49 & \underline{77.09} \\
\ding{51} & \ding{51} & - & \ding{51} & 92.00 & 68.45 & 92.95 & 74.38 & 93.35 & 75.85 \\
\ding{51} & - & \ding{51} & \ding{51} & 90.24 & 65.30 & 92.29 & 72.18 & 92.88 & 74.73 \\
- & \ding{51} & \ding{51} & \ding{51} & 91.37 & 65.31 & 93.07 & 73.16 & \textbf{93.83} & \textbf{79.06} \\	
\ding{51} & \ding{51} & \ding{51} & \ding{51} & 91.39 & 69.83 & 92.25 & 70.81 & \underline{93.61} & 75.59 \\
\bottomrule
\end{tabular*}
\label{tab:aug_ablation}
\end{table}

\begin{table}[t!]
\setlength\tabcolsep{0pt}
 \centering
 \small
 \caption{Finetuning results across different training sizes for three Canaries. SSL pretraining and semi-supervised post-training (Post) are consistently effective. Moreover, the data augmentations (Aug) are particularly useful for few-shot settings.}
 \begin{tabular*}{\textwidth}{@{\extracolsep{\fill}} l*{10}{c}}
    \toprule
    \multirow{2}{*}{Bird} & \multirow{2}{*}{Init} & \multirow{2}{*}{Aug} & \multirow{2}{*}{Post} & \multicolumn{2}{c}{Few-shot} & \multicolumn{2}{c}{Few-shot + 1\%} & \multicolumn{2}{c}{Few-shot + 2\%}\\
    \cline{5-6} \cline{7-8} \cline{9-10}
    & & & & Acc & F1 & Acc & F1 & Acc & F1\\
    \midrule   
     \multirow{7}{*}{llb3}& Random   & -         & -         & 89.35 & 62.36 & 95.94 & 86.16 & 96.36 & 87.73 \\
                          & MAE      & -         & -         & 92.31 & 72.20 & \textbf{96.35} & \textbf{88.39} & 96.61 & 89.44 \\
                          & MAE      & \ding{51} & -         & 93.53 & 77.57 & 96.23 & 87.80 & 96.78 & 90.05 \\
                          & MAE      & \ding{51} & \ding{51} & \textbf{94.09} & \textbf{79.96} & \underline{96.33} & \underline{88.25} & \textbf{96.89} & \textbf{90.42} \\
                          & OSC  & -         & -         & 92.44 & 72.55 & 96.21 & 87.67 & 96.58 & 89.02 \\
                          & OSC  & \ding{51} & -         & 93.49 & 77.09 & 96.28 & 88.05 & 96.78 & 89.74 \\
                          & OSC  & \ding{51} & \ding{51} & \underline{93.96} & \underline{79.50} & 96.28 & 88.11 & \underline{96.85} & \underline{90.10} \\
    \midrule
    \multirow{7}{*}{llb11}& Random  & -         & -         & 93.24 & 68.52 & 98.08 & 90.22 & 98.31 & 92.09 \\
                           & MAE     & -         & -         & 95.40 & 79.82 & 98.35 & 93.15 & 98.47 & \underline{94.01} \\
                           & MAE     & \ding{51} & -         & 96.54 & 83.19 & 98.35 & 93.02 & 98.41 & 93.67 \\
                           & MAE     & \ding{51} & \ding{51} & \underline{96.94} & \textbf{86.09} & \textbf{98.43} & \underline{93.33} & \underline{98.50} & \underline{94.01} \\
                           & OSC & -         & -         & 96.61 & 84.71 & \underline{98.36} & 92.84 & 98.42 & 93.10 \\
                           & OSC & \ding{51} & -         & 96.69 & 84.13 & 98.35 & 93.15 & 98.44 & 93.54 \\
                           & OSC & \ding{51} & \ding{51} & \textbf{96.96} & \underline{85.80} & \textbf{98.43} & \textbf{93.59} & \textbf{98.53} & \textbf{94.02} \\
    \midrule
    \multirow{7}{*}{llb16}& Random  & -         & -         & 94.15 & 70.48 & 96.97 & 87.24 & 97.02 & 86.81 \\
                           & MAE     & -         & -         & 95.72 & 80.88 & 97.30 & 89.91 & 97.33 & 89.29 \\
                           & MAE     & \ding{51} & -         & \underline{95.80} & 82.55 & 97.40 & 90.23 & 97.41 & 90.13 \\
                           & MAE     & \ding{51} & \ding{51} & \textbf{96.02} & \textbf{84.04} & \textbf{97.47} & 90.18 & \underline{97.50} & 90.28 \\
                           & OSC & -         & -         & 95.17 & 81.34 & \underline{97.42} & \underline{90.93} & 97.43 & 90.31 \\
                           & OSC & \ding{51} & -         & 95.22 & 81.15 & 97.37 & 90.59 & 97.43 & \underline{90.89} \\
                           & OSC & \ding{51} & \ding{51} & 95.49 & \underline{82.97} & \textbf{97.47} & \textbf{91.13} & \textbf{97.51} & \textbf{90.92} \\
    \bottomrule
 \end{tabular*}
 \label{tab:canary_results}
 % \vspace{-10pt}
\end{table}

\subsection{Data Augmentation Ablation Study}
Augmentation techniques do not necessarily have a positive compounding effect, and the proper setup depends on the dataset, task, and model~\cite{NANNI2020101084,gontijo2020affinity,steiner2022how}. \autoref{tab:aug_ablation} shows the augmentation ablation in this work. The purpose of this experiment is to estimate the potential utility of multiple promising techniques for birdsong syllable detection before conducting the complete finetuning experiments. The ablation used the few-shot training set from one of the birds (llb3) to prevent information leakage from the test sets into subsequent experiments. Moreover, ablation was conducted using three model initializations: random, MAE pretraining, and OSC pretraining. These models are finetuned end-to-end (i.e., no encoder freezing). Although Time-Frequency Masking could be effective, combining random gain, color noise, and Bernoulli noise has an overall advantage. Therefore, the next experiments use this combination.

\subsection{Evaluating the Proposed Three-Stage Training Pipeline}
\autoref{tab:canary_results} shows the main experiment that comprehensively evaluates the proposed method. The exact values of these scores vary in practice with hyperparameters and datasets, and this work is not an incremental improvement on a public benchmark. What matters here is to gain insight. First, note that both SSL-pretrained models consistently outperform random initialization across training set sizes and individuals. Additionally, the data augmentation is particularly useful in the few-shot setting. However, the performance seems to reach a ceiling as the training size grows (with negligible fluctuation), which is due to the peculiar distribution of a bird’s syllables, i.e., extreme class imbalance. This indicates that the number of recordings or duration is not an appropriate measure of required training data. This will be discussed further, along with practical guidelines, applications, and future directions. A similar argument holds for the Semi-SL pot-training. Overall, the training framework is robust, even without a validation set for tuning and early stopping.

\begin{table}
\setlength\tabcolsep{0pt}
 \centering
 \small
 \caption{The columns show the number of recording files and their total duration in minutes (m) in each split.}
 \begin{tabular*}{\textwidth}{@{\extracolsep{\fill}} l*{4}{c}}
    \toprule
    Bird & Train & Validation & Test  & Syllables\\
    \midrule
    bl26lb16 & 8 (3 m) & 17 (7 m) & 153 (62 m) & 7\\
    gr41rd51 & 38 (9 m) & 77 (17 m) & 656 (143 m) & 11\\
    gy6or6   & 39 (7 m) & 78 (14 m) & 663 (118 m) & 11\\
    or60yw70 & 16 (4 m) & 33 (8 m) & 289 (66 m) & 8\\
    \bottomrule
 \end{tabular*}
 \label{tab:data_bf}
\end{table}

\section{Methodological Generalization: Case Study of Bengalese Finch}
This section demonstrates the general applicability of the Res-MLP-RNN, SSL algorithms, and training pipeline for another species, without any hyperparameter tuning. This validates the claim that a successful method for the Canary will establish a strong baseline.

The dataset consists of laboratory recordings from four Bengalese finches, open-sourced by Nicholson~et~al.~\cite{Nicholson2022}. They are sampled at 32~kHz and fully annotated at the syllable level. The dataset is split into train ($5\%$), validation ($10\%$), and test ($85\%$) sets. The split is done such that at least one example of each syllable class is present in all three sets. \autoref{tab:data_bf} shows the details of the dataset. The data preprocessing, SSL pretraining, and the rest of the training pipeline follow the exact same procedure described for the Canary. The validation set was primarily used to monitor the Semi-SL stage, as its success is task and data-dependent. Previous research has demonstrated that Semi-SL methods are sensitive to data distributions and hyperparameter configurations and may degrade performance below that of a standard supervised baseline if not used with care \cite{li2016towards,NEURIPS2018_c1fea270}.

\autoref{tab:bf_results} shows the results. Note that the exact values of these scores are not important here, as they vary slightly with augmentation and hyperparameters. As before, the key point is that both SSL pretraining methods surpass random initialization, indicating that the models learned meaningful feature representations of the Bengalese Finch song.

\begin{table}[t!]
\setlength\tabcolsep{0pt}
 \centering
 \footnotesize
 \caption{Finetuning results for Bengalese Finches. The results confirm the findings from the Canary experiments, which had more difficult songs than those of Bengalese finches.}
 \begin{tabular*}{\textwidth}{@{\extracolsep{\fill}} l*{5}{c}}
    \toprule
    Bird & Initialization & Augmentation & Post-training & Accuracy & F1-Macro \\
    \midrule   
     \multirow{7}{*}{bl26lb16}& Random   & -         & -         & 98.10 & 95.34 \\
                          & MAE      & -         & -         & 98.42 & 96.73 \\
                          & MAE      & \ding{51} & -         & 98.43 & 96.69 \\
                          & MAE      & \ding{51} & \ding{51} & \textbf{98.60} & \textbf{96.95} \\
                          & OSC  & -         & -         & \underline{98.59} & 96.46 \\
                          & OSC  & \ding{51} & -         & 98.40 & 96.62 \\
                          & OSC  & \ding{51} & \ding{51} & 98.47 & \underline{96.80} \\
    \midrule

    \multirow{7}{*}{gr41rd51}& Random   & -         & -      & 98.31 & 96.63 \\
                          & MAE      & -         & -         & 98.72 & 97.61 \\
                          & MAE      & \ding{51} & -         & 98.68 & 97.57 \\
                          & MAE      & \ding{51} & \ding{51} & \underline{98.76} &  \underline{97.67} \\
                          & OSC  & -         & -         & 98.75 & 97.66 \\
                          & OSC  & \ding{51} & -         & 98.67 &  97.59 \\
                          & OSC  & \ding{51} & \ding{51} & \textbf{98.79} & \textbf{97.74} \\
    \midrule

    \multirow{7}{*}{gy6or6}& Random   & -         & -         & 98.26 & 97.86 \\
                          & MAE      & -         & -         & \textbf{98.61} & 98.36 \\
                          & MAE      & \ding{51} & -         & 98.53 & 98.28 \\
                          & MAE      & \ding{51} & \ding{51} & 98.59 & 98.34 \\
                          & OSC  & -         & -         & 98.55 & 98.31 \\
                          & OSC  & \ding{51} & -         & 98.57 & \underline{98.35} \\
                          & OSC  & \ding{51} & \ding{51} & \underline{98.60} & \textbf{98.38} \\
    \midrule

    \multirow{7}{*}{or60yw70}& Random   & -         & -      & 98.42 & 97.78 \\
                          & MAE      & -         & -         & 98.64 & 98.20 \\
                          & MAE      & \ding{51} & -         & 98.72 & 98.30 \\
                          & MAE      & \ding{51} & \ding{51} & 98.72 & 98.30 \\
                          & OSC  & -         & -         & 98.73 & 98.32 \\
                          & OSC  & \ding{51} & -         & \underline{98.75} & \underline{98.32} \\
                          & OSC  & \ding{51} & \ding{51} & \textbf{98.80} & \textbf{98.40} \\
    \bottomrule

 \end{tabular*}
 \label{tab:bf_results}
\end{table}

\begin{figure}[t]
     \centering
     \begin{subfigure}[t]{\textwidth}
         \centering
         \includegraphics[width=\textwidth]{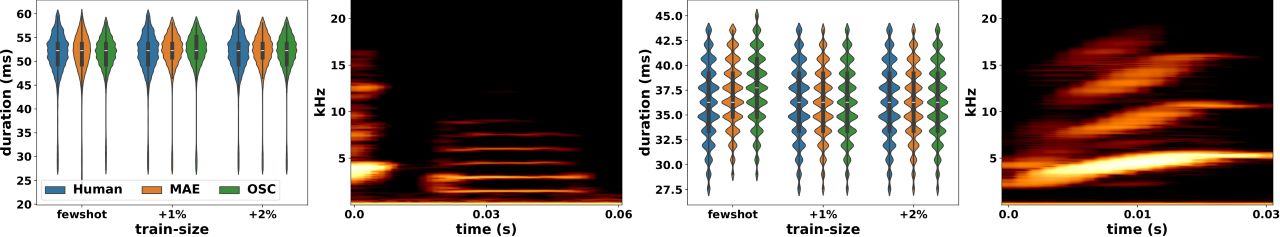}
         \caption{llb3}
         % \label{fig:llb3_syllable_duration}
     \end{subfigure}
     \par\medskip
     \begin{subfigure}[t]{\textwidth}
         \centering
         \includegraphics[width=\textwidth]{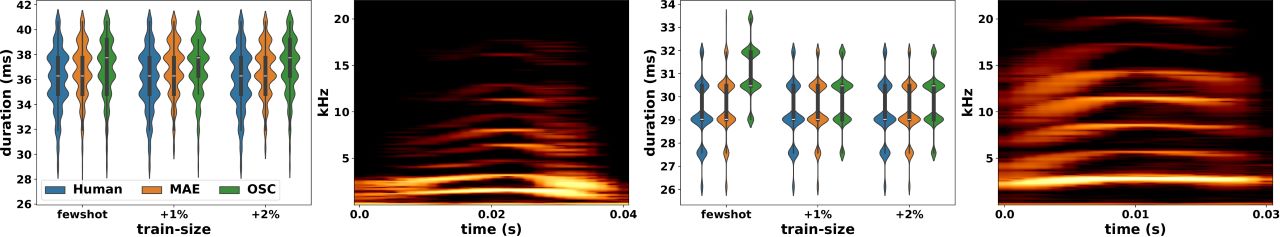}
         \caption{llb11}
     \end{subfigure}
     \par\medskip
     \begin{subfigure}[t]{\textwidth}
         \centering
         \includegraphics[width=\textwidth]{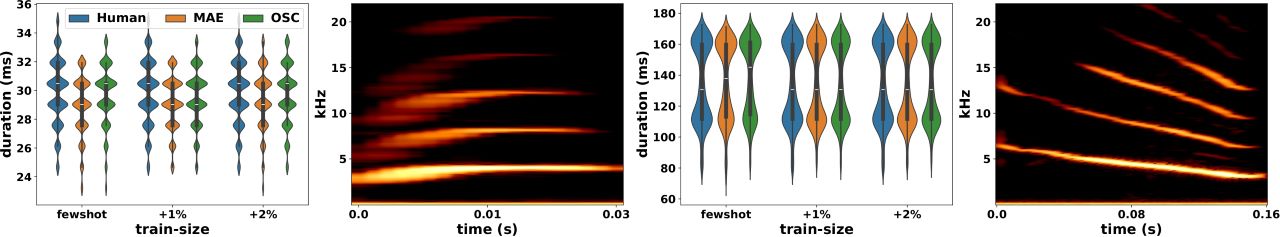}
         \caption{llb16}
     \end{subfigure}
    \caption{Two syllables from each bird with their true and predicted distribution of duration across training sizes and pretrained models. The predictions are faithful.}
    \label{fig:durations}
    % \vspace{0pt}
\end{figure}

\begin{figure}[t!]
% \vspace{0pt}
     \centering
     \begin{subfigure}[t]{.24\textwidth}
         \centering
         \includegraphics[width=\textwidth]{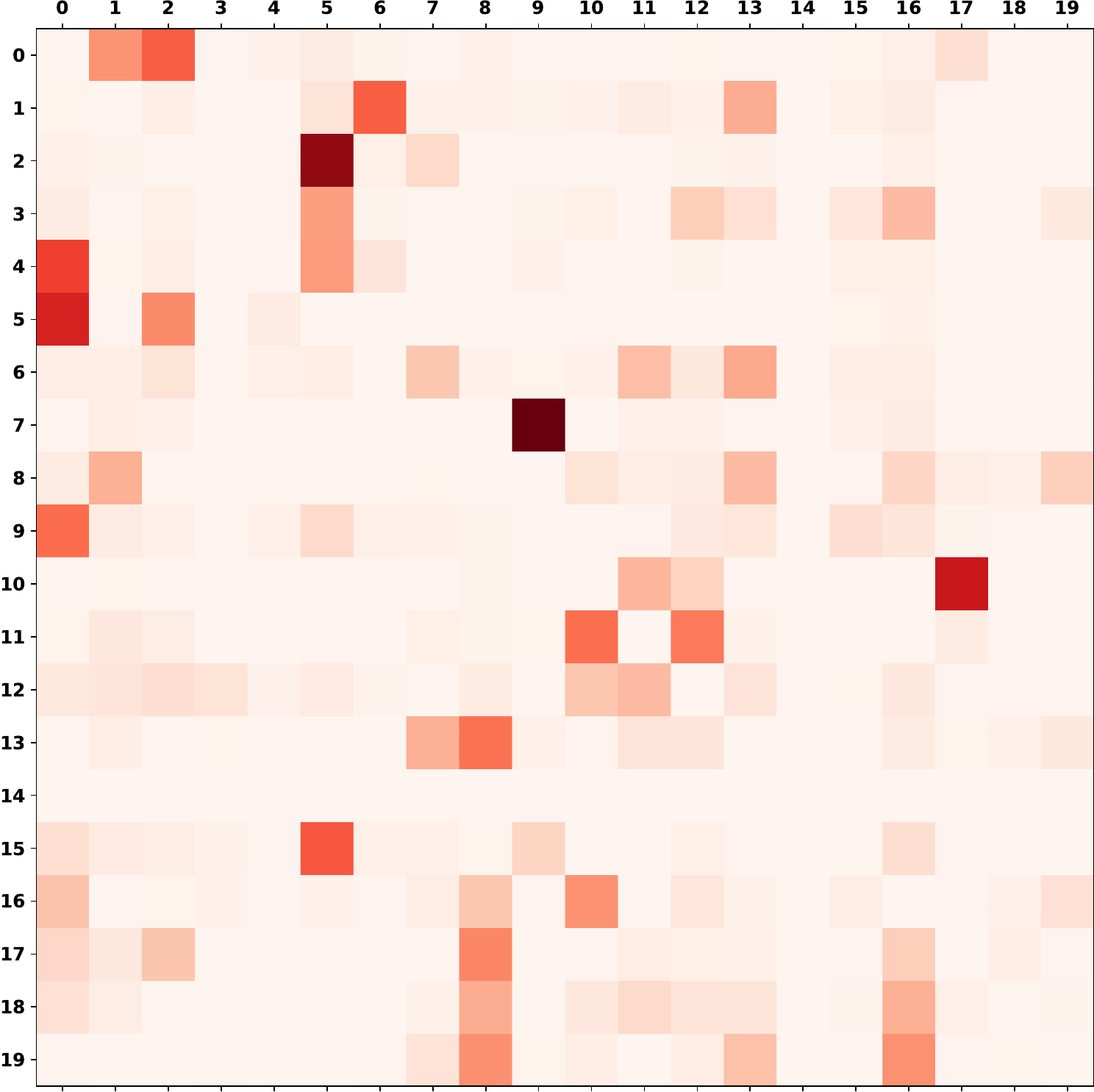}
         \caption{llb3, MAE, few-shot}
         % \label{fig:x}
     \end{subfigure}
     \hfill
     \begin{subfigure}[t]{.24\textwidth}
         \centering
         \includegraphics[width=\textwidth]{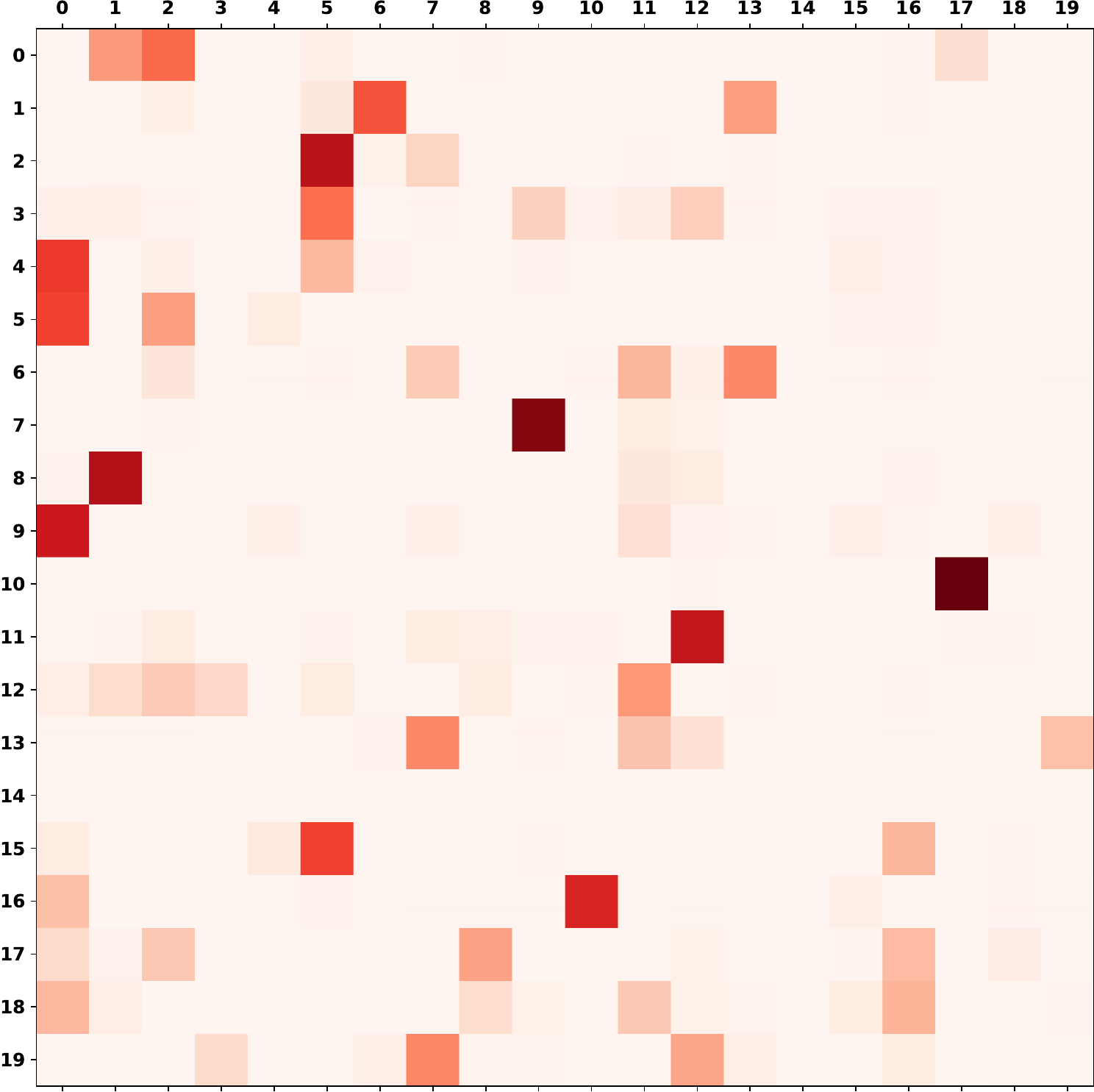}
         \caption{llb3, MAE, +1\%}
     \end{subfigure}
     \hfill
     \begin{subfigure}[t]{.24\textwidth}
         \centering
         \includegraphics[width=\textwidth]{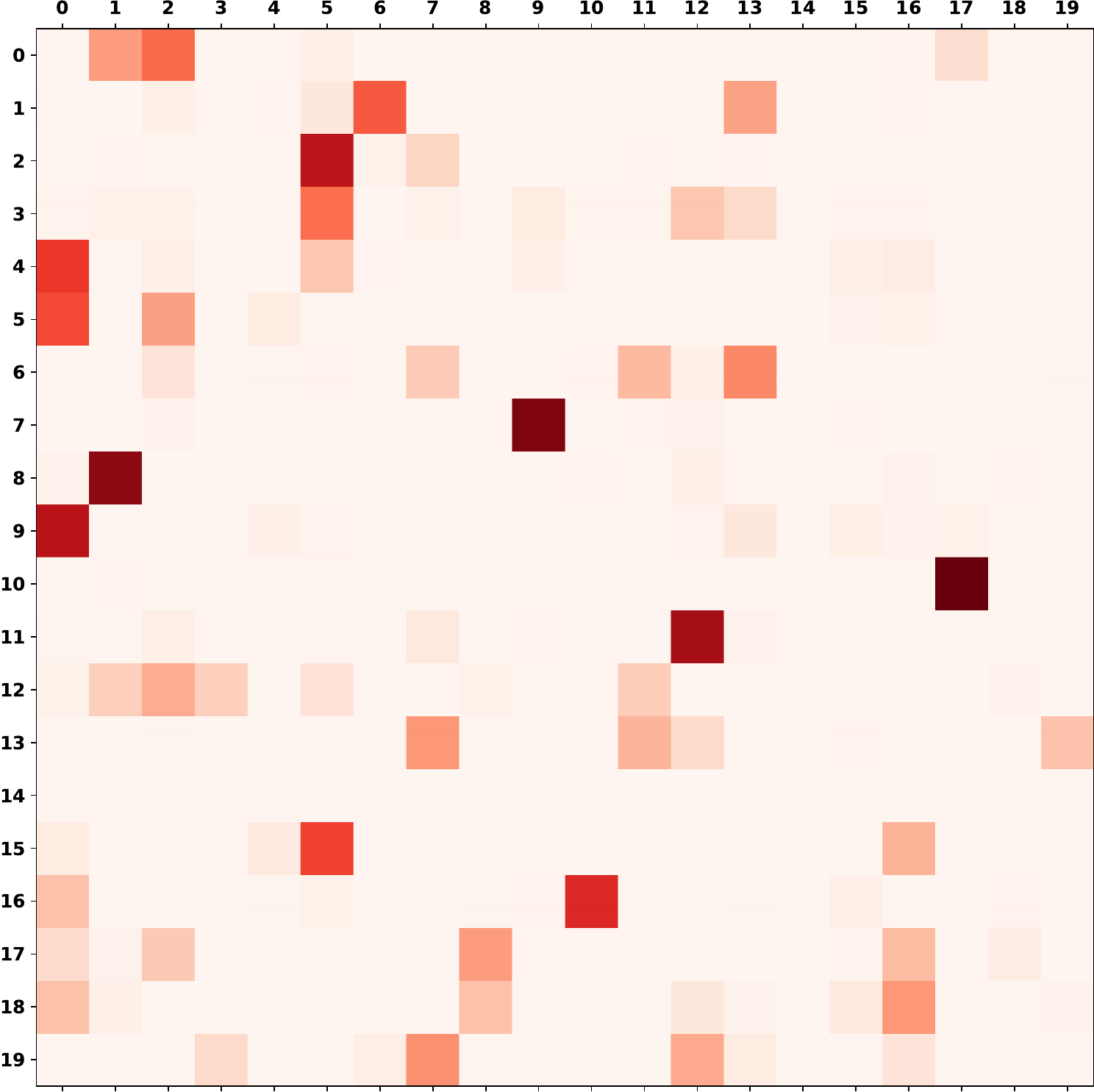}
         \caption{llb3, MAE, +2\%}
     \end{subfigure}
     \hfill
     \begin{subfigure}[t]{.24\textwidth}
         \centering
         \includegraphics[width=\textwidth]{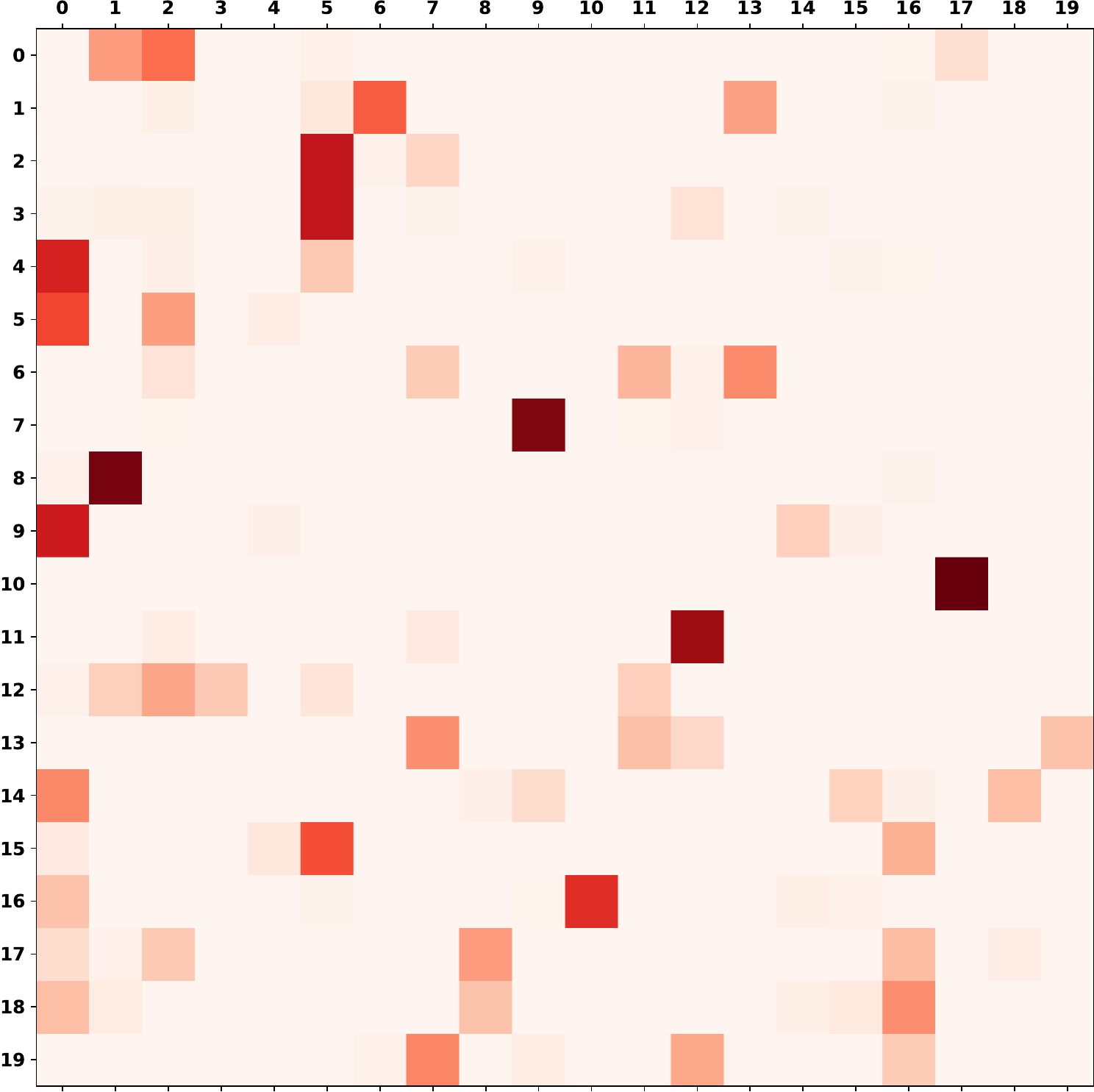}
         \caption{llb3, True Matrix}
     \end{subfigure}
     \par\medskip
     \begin{subfigure}[t]{.24\textwidth}
         \centering
         \includegraphics[width=\textwidth]{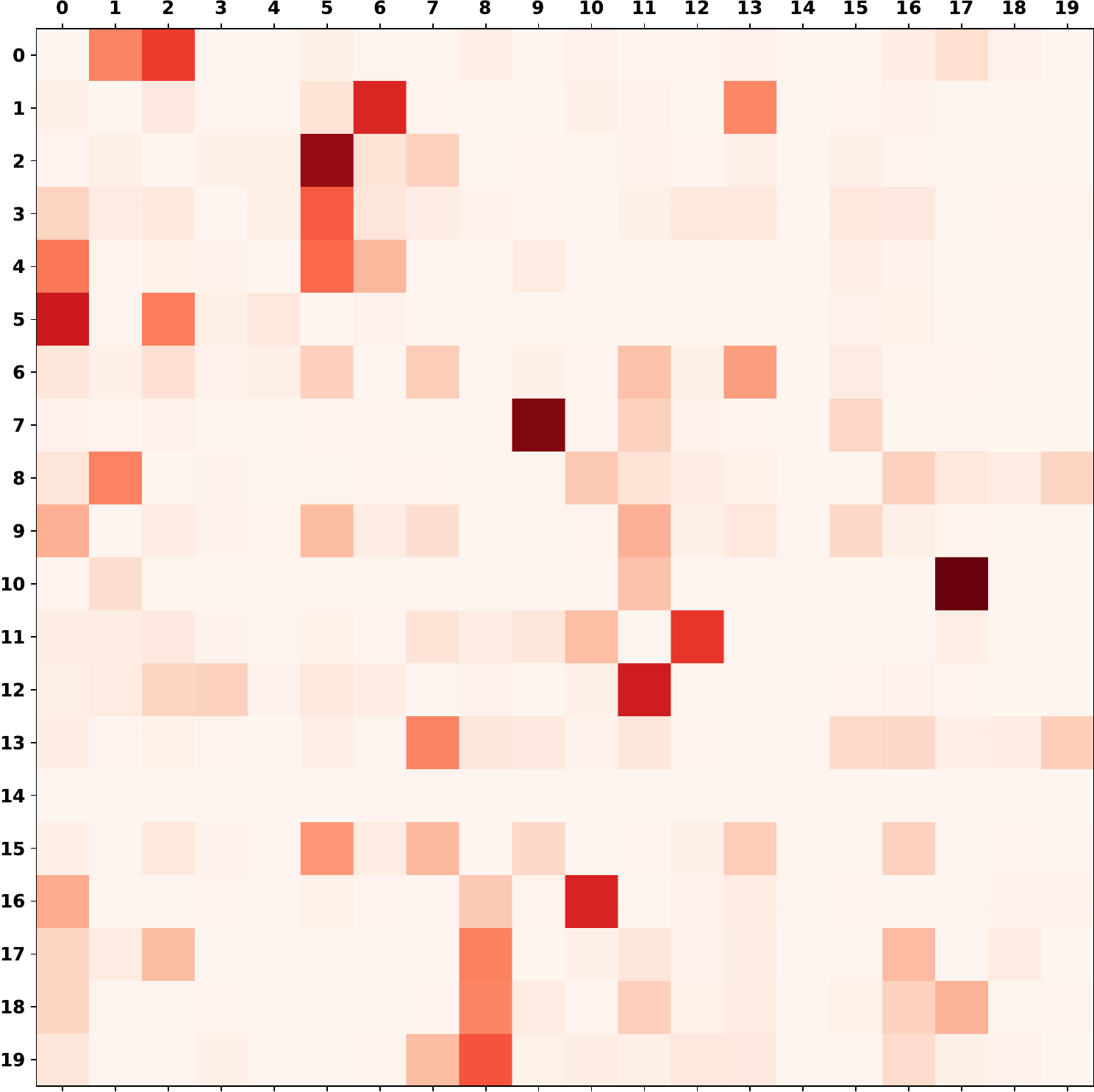}
         \caption{llb3, OSC, few-shot}
         % \label{fig:x}
     \end{subfigure}
     \hfill
     \begin{subfigure}[t]{.24\textwidth}
         \centering
         \includegraphics[width=\textwidth]{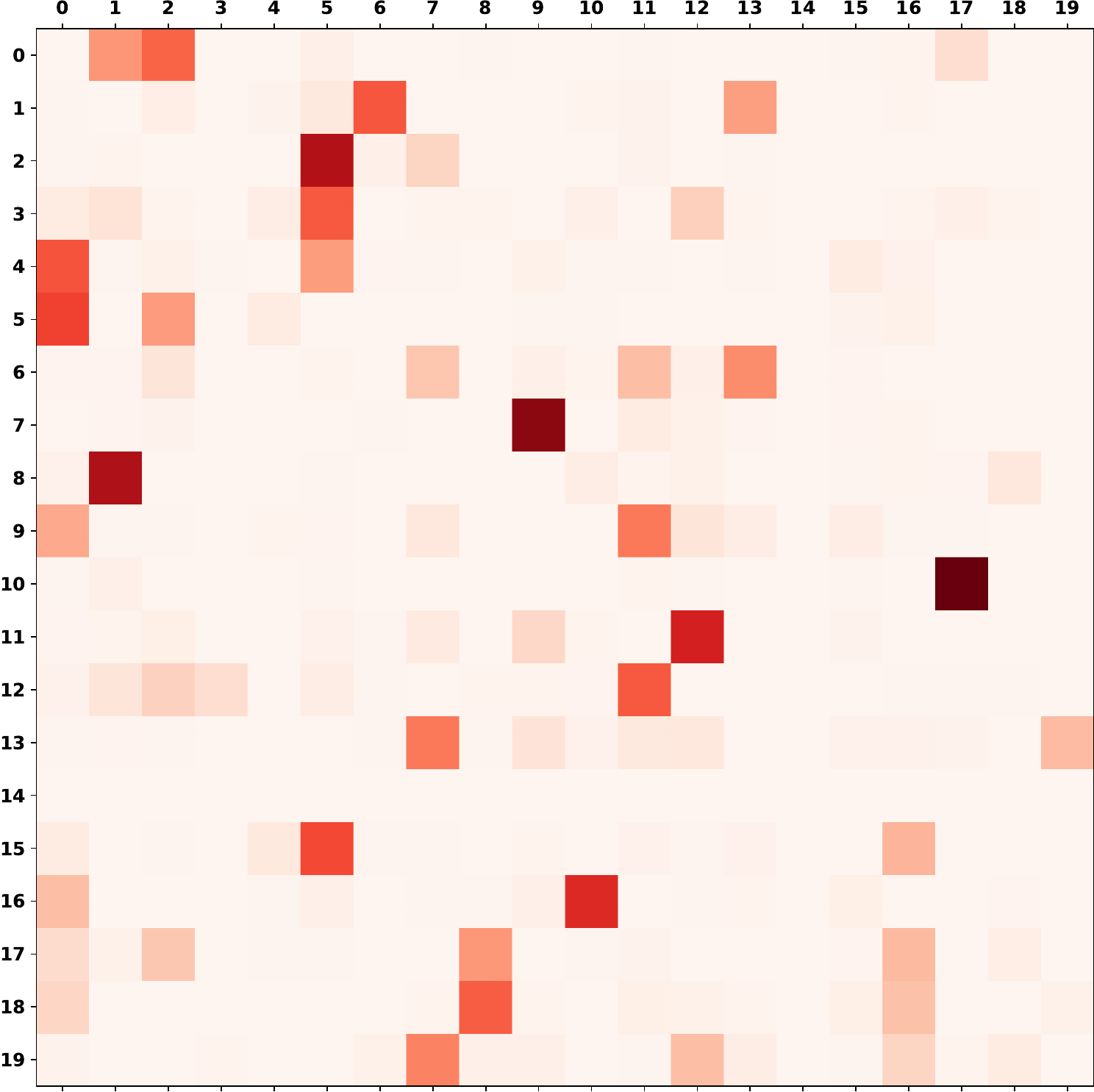}
         \caption{llb3, OSC, +1\%}
     \end{subfigure}
     \hfill
     \begin{subfigure}[t]{.24\textwidth}
         \centering
         \includegraphics[width=\textwidth]{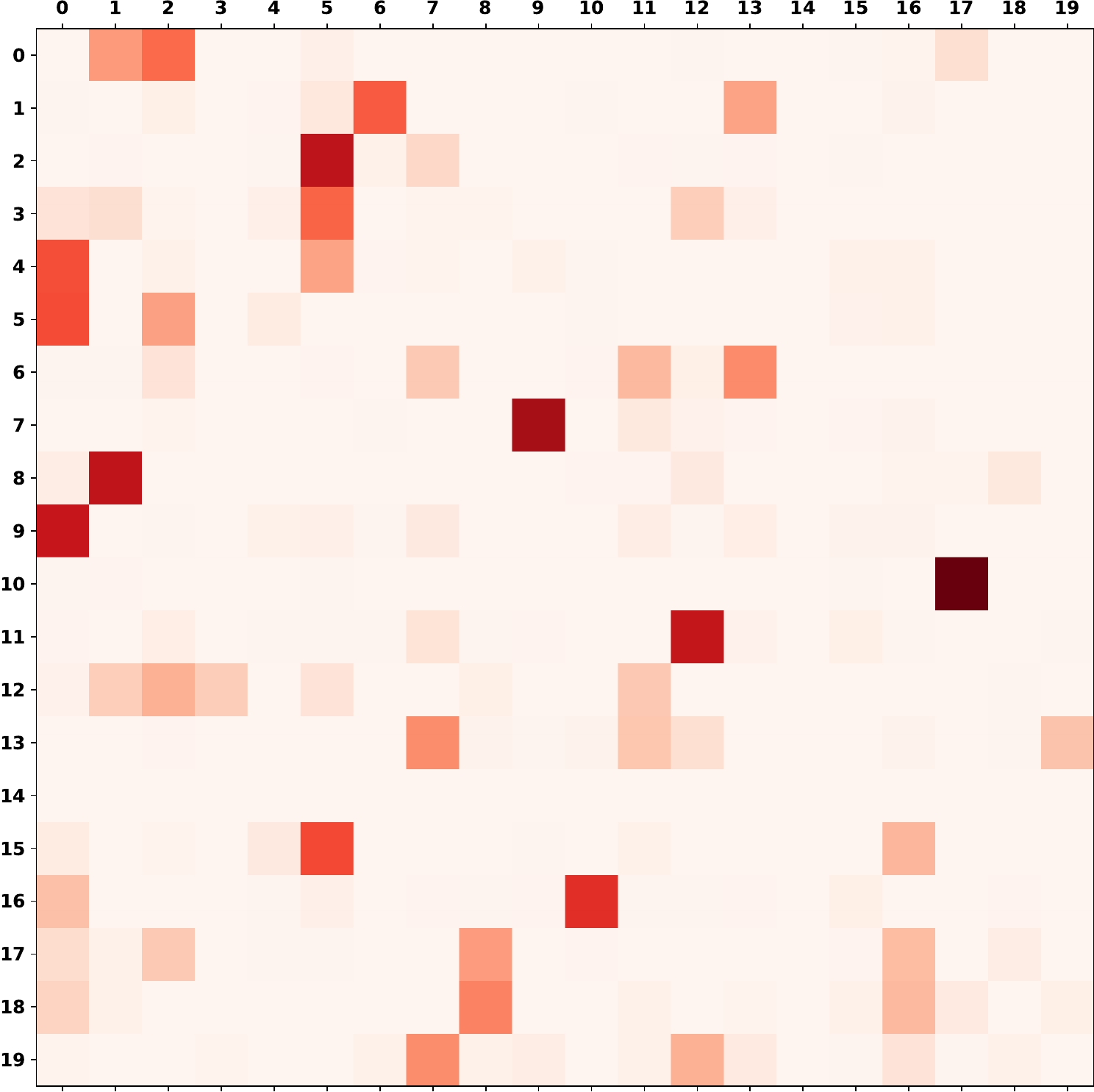}
         \caption{llb3, OSC, +2\%}
     \end{subfigure}
     \hfill
     \begin{subfigure}[t]{.24\textwidth}
         \centering
         \includegraphics[width=\textwidth]{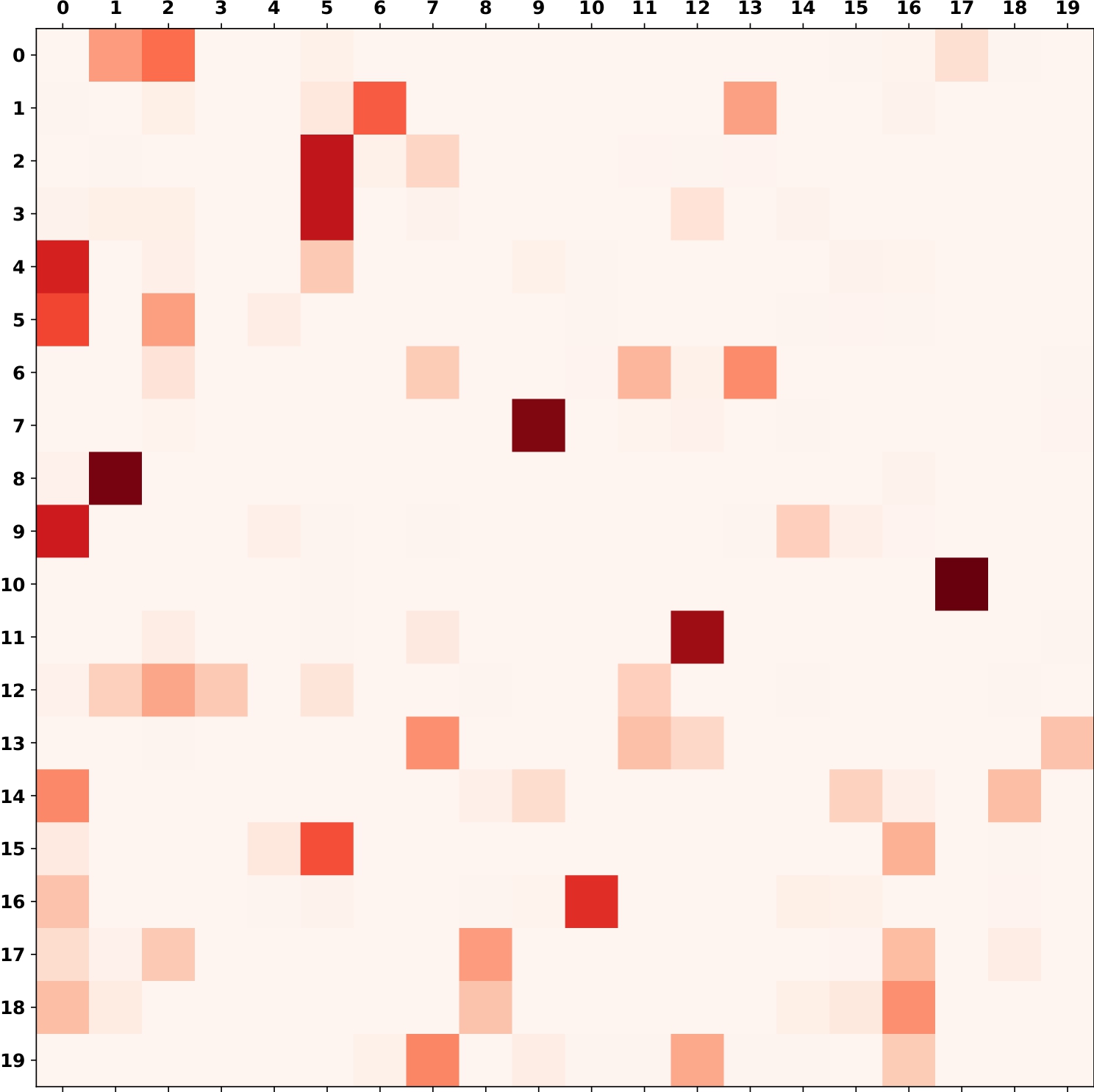}
         \caption{llb3, True Matrix}
     \end{subfigure}
    \caption{Syllable transition matrices for the llb3 canary across pretrained models and data sizes. Even with few-shot finetuning, the structure of the transition is visible. The True Matrix is identical across rows and is computed from human annotations. Each cell is a syllable pair, where the diagonal of the matrix is a syllable transition to itself.}
    \label{fig:tms}
\end{figure}

% \subsection{Assessing Multi-Species SSL Embeddings via Linear Probing}
% mae llb3 1 => acc: 94.4, f1: 83.54, precision: 87.79, recall: 80.5
% osc llb3 1 => acc: 95.75, f1: 87.85, precision: 88.08, recall: 87.8
% mae llb11 1 => acc: 95.87, f1: 78.79, precision: 89.11, recall: 75.0
% osc llb11 1 => acc: 97.75, f1: 90.8, precision: 93.03, recall: 90.17
% mae llb16 1 => acc: 94.78, f1: 74.17, precision: 87.0, recall: 70.44
% osc llb16 1 => acc: 96.11, f1: 86.24, precision: 89.91, recall: 84.65

\section{Applications and Future Directions}\label{sec:applications}
This section investigates how reliably the developed models can extract informative statistics about the structure of birdsong for downstream studies. This also helps estimate the required amount of data to develop trustworthy models. Additionally, an unsupervised analysis is discussed, which can be used for data curation and other purposes in future studies.

\textbf{Syllable duration}: One important factor of birdsong is the duration of the vocalized syllables, as they vary from rendition to rendition. \autoref{fig:durations} shows the true and predicted distribution of duration for some syllables across models and training sizes. The results are faithful, even in the few-shot.

\textbf{Birdsong syntax}: Another important aspect of the birdsong is the transition of syllables. The sequences of vocalized syllables in birdsongs sometimes form sophisticated and probabilistic branching patterns with long-range context-dependency~\cite{morita2021measuring,cohen2022automated}. As a simplified probe, it would be interesting to study the syllable transition patterns using bigrams, i.e., the frequency of immediate transitions from a particular syllable to others~\cite{berwick2011songs}. As illustrated in \autoref{fig:tms}, it provides insight into the grammar of a bird's songs.

\textbf{SSL applications}: We assess the quality of SSL embeddings by clustering them and evaluating their utility. First, the syllables are isolated based on true temporal boundaries. Each syllable is fed into the SSL models, and the embeddings are reduced to 32 principal components; the syllable duration is appended as an additional feature. The features are standardized and clustered with a Gaussian Mixture Model. The number of clusters is the true number of syllables for each bird to enable direct comparison with annotations. \autoref{tab:clustering_ami} shows the Adjusted Mutual Information (AMI)~\cite{vinh2009information} scores for each canary and SSL model.
\begin{wraptable}{R}{0.35\textwidth}
% \vspace{-25pt}
\setlength\tabcolsep{0pt}
 \centering
 \scriptsize
 \caption{Clustering results.}
 \begin{tabular*}{0.35\textwidth}{@{\extracolsep{\fill}} l*{2}{c}}
    \toprule
    Bird & Embeddings & AMI \\
    \midrule
    \multirow{2}{*}{llb3} & MAE &  0.694\\
    & OSC & 0.636\\
    \midrule
    \multirow{2}{*}{llb11} & MAE & 0.804\\
    & OSC & 0.755\\
    \midrule
    \multirow{2}{*}{llb16} & MAE & 0.762\\
    & OSC & 0.732\\
    \bottomrule
 \end{tabular*}
 \label{tab:clustering_ami}
 % \vspace{-20pt}
\end{wraptable}

\begin{figure}[t]
     \centering
     \begin{subfigure}[t]{0.49\textwidth}
         \centering
         \includegraphics[width=\textwidth]{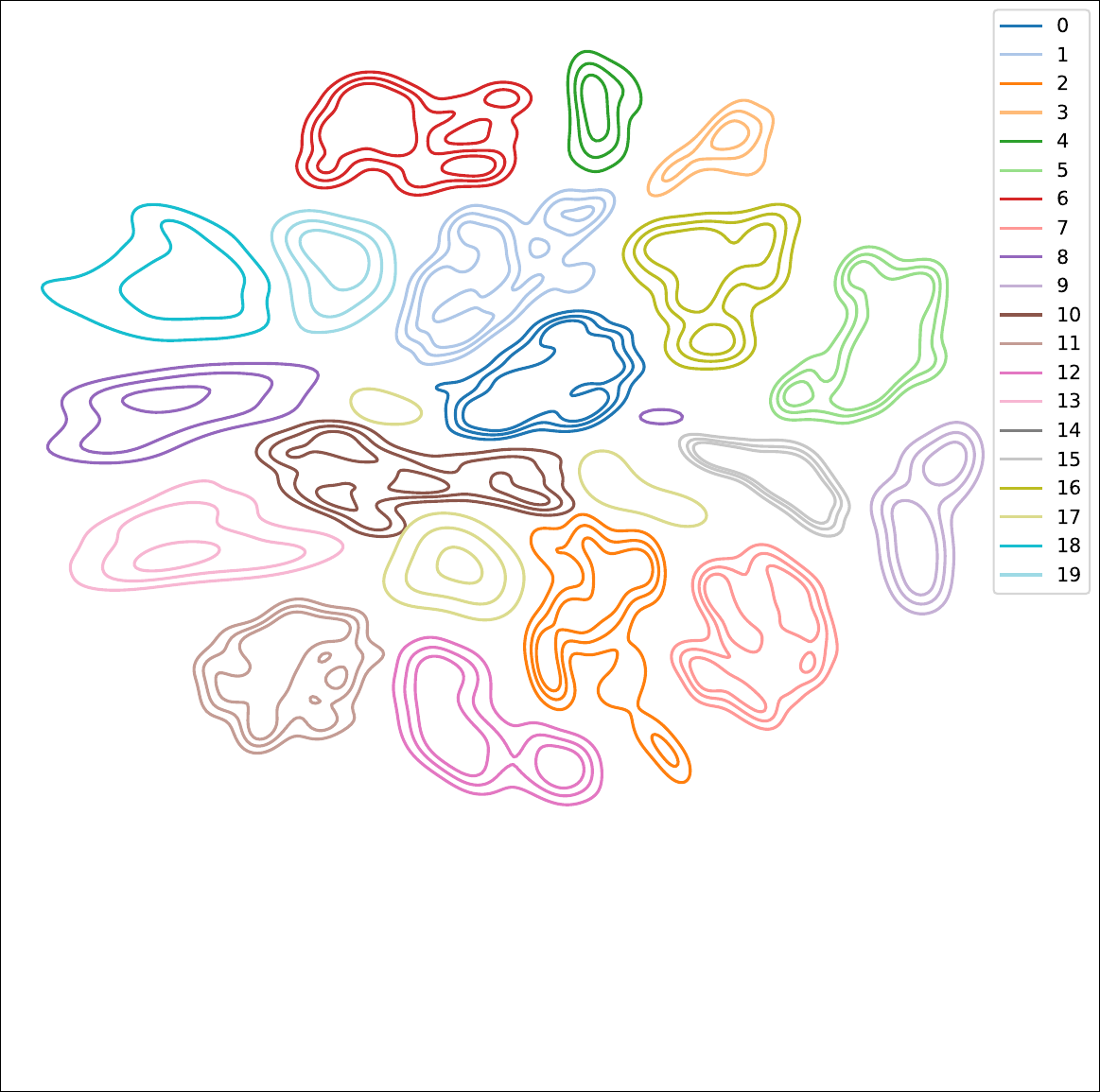}
         \caption{llb3, MAE}
     \end{subfigure}
     \hfill
     \begin{subfigure}[t]{0.49\textwidth}
         \centering
         \includegraphics[width=\textwidth]{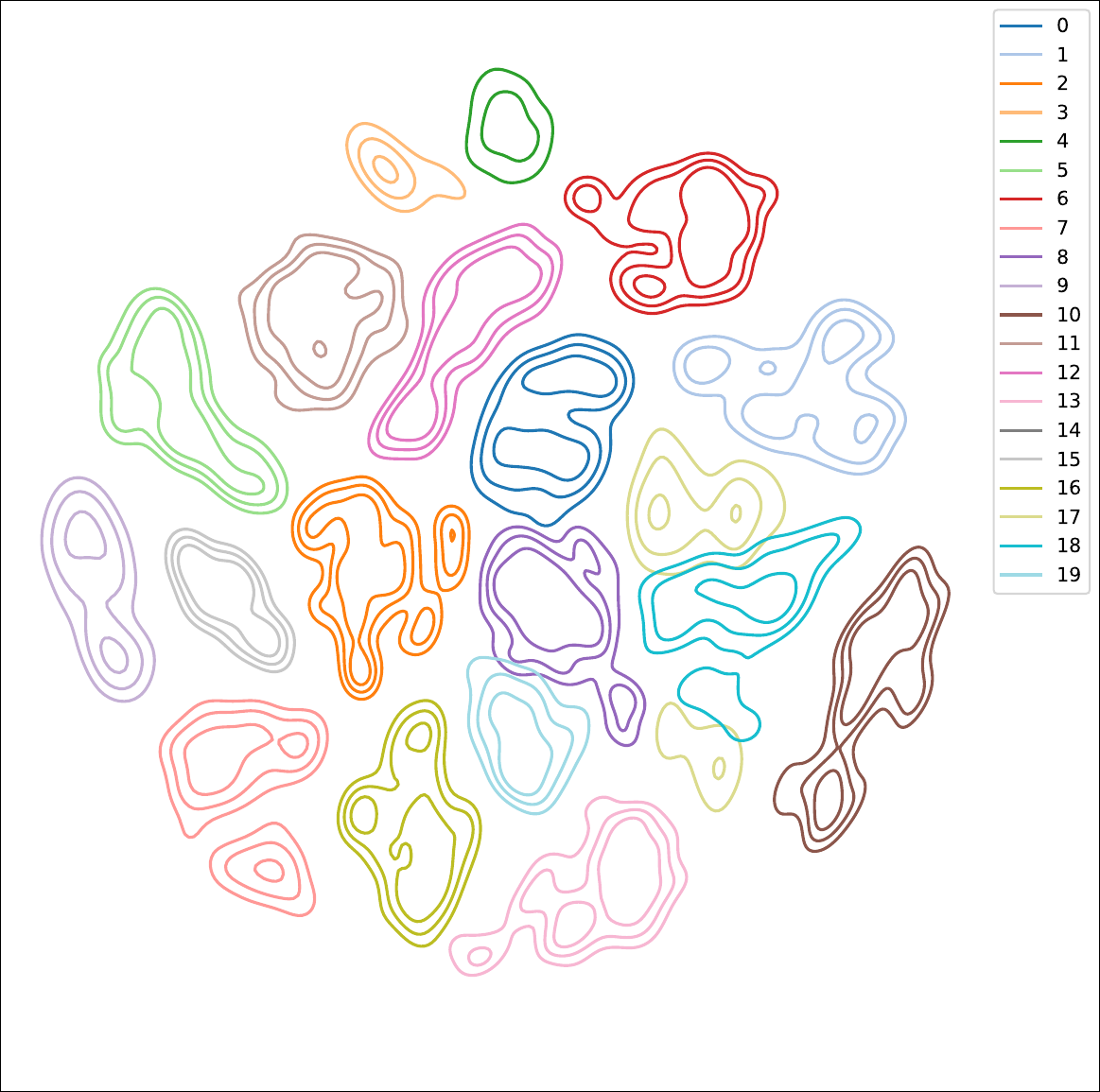}
         \caption{llb3, OSC}
     \end{subfigure}
     \caption{T-SNE plots of the clustered SSL embeddings of syllables after re-labeling via majority vote. To avoid clutter, the plots show the density contours for each cluster, estimated from a maximum of 2000 randomly sampled syllables.}
    \label{fig:tsne}
    % \vspace{-15pt}
\end{figure}

Next, the syllables within each cluster are re-labeled with a true syllable label via majority vote. \autoref{fig:tsne} shows the T-SNE~\cite{van2008visualizing} plots of the embeddings for syllable clusters after re-labeling. It is remarkably well-organized despite the lack of human supervision in SSL.

The true syllable segmentation and labels were used for clarity. An application of this procedure is data curation~\cite{rauch2024deepactivelearningavian}. There is no need to know the true number of syllables, which is partly subjective to begin with. In fact, over-clustering would benefit the data-curation process. For syllables' temporal boundary segmentation, a semi-supervised method was proposed by Ghaffari~and~Devos~\cite{houtanfa} that yields near-expert results with only a few seconds of inexpensive binary annotation. Moreover, they used a CRNN with random initialization, and our Res-MLP-RNN and SSL pretraining can further improve this segmentation. Due to extreme class imbalance, as evident by the saturated F1 in our results, the number of recordings or their duration is not a precise measure of the effective amount of the training data for birdsong. The experts can use an SSL model to perform a clustering analysis and identify recordings with potentially different syllable patterns. Afterward, a small but highly effective training set can be curated by labeling a few recordings from each cluster. Approximately 10 minutes of well-curated data appears sufficient to train extremely reliable models.

\textbf{Future directions}: SSL models, if pretrained on sufficiently large datasets, can be made open-source for off-the-shelf use. Beyond data curation, improving SSL syllable embeddings can enable novel, more precise studies of birdsong without labels. For example, measuring song similarity is an important research area~\cite{tchernichovski2021balanced}. It is debatable whether publicly available outdoor bird datasets, such as BirdSet~\cite{rauch2025_birdset}, are suitable for the SSL stage of fine-grained birdsong analysis. This requires validation by experts in downstream biological tasks, but it is an interesting topic for future study. Ultimately, the birdsong analysis could be extended beyond individual-specific (or even species-specific) models and largely eliminate annotation labor by applying SSL methods, combined with improved segmentation and clustering from prior work~\cite{houtanfa,morita2021measuring}.

\section{Conclusion}\label{sec:conclusion}
This work alleviates a challenge at the intersection of computational bioacoustics and related biological disciplines that rely on birdsong analysis: the substantial expert labor required for detailed, syllable-level annotation. To this end, we proposed the Res-MLP-RNN, a data-efficient neural network architecture that effectively captures the complex temporal and harmonic structures of birdsongs. Furthermore, a comprehensive three-stage training framework was introduced, comprising self-supervised pretraining, supervised fine-tuning, and semi-supervised post-training. Notably, two established self-supervised paradigms, masked prediction and online clustering, were adapted for birdsong analysis.

We validated the proposed framework on two species with distinct vocal complexities, the Canary and the Bengalese Finch. The results confirm effective generalization across species without requiring large pretraining datasets or hyperparameter tuning for the Res-MLP-RNN and self-supervised algorithms. Given the need for rapid model development with limited per-bird data in syllable annotation tasks, Res-MLP-RNN achieved fast convergence and robust predictions even with minimal training data across various model sizes. These capabilities distinguish the presented model and algorithms from data-intensive Transformer-based self-supervised models, which are highly sensitive to hyperparameters.

Beyond algorithmic efficiency, the framework offers practical tools for researchers. Reducing the annotation workload to a few minutes of audio enables experts to focus on downstream linguistic, behavioral, and neurological analyses. The robust feature representations from self-supervised models also support unsupervised birdsong analysis, including automated data curation and cross-birdsong similarity assessment. Integrating these methods into biological workflows will improve the reproducibility, scalability, and scope of future research.

\bibliographystyle{elsarticle-num}
\bibliography{ref}




% %% For numbered reference style

%   \textit{\LaTeX: a document preparation system},
%   Addison Wesley, Massachusetts,

% \end{thebibliography}
\end{document}